\documentclass[11pt]{article}

\usepackage[preprint]{acl}

\usepackage{times}
\usepackage{latexsym}
\usepackage{enumitem}
\usepackage{amsmath}
\usepackage{algorithm}
\usepackage{algpseudocode}
\usepackage{amsmath}
\usepackage{pifont}
\usepackage{xcolor}
\usepackage{tcolorbox}
\providecommand{\xmark}{\ding{55}}
\providecommand{\xmark}{\ding{55}}

\usepackage[T1]{fontenc}

\usepackage[utf8]{inputenc}

\usepackage{microtype}

\usepackage{inconsolata}

\usepackage{graphicx}

\usepackage{tcolorbox}
\usepackage{booktabs}
\usepackage{multirow}

\title{MetaGAI: A Large-Scale and High-Quality Benchmark for Generative AI Model and Data Card Generation}

\author{
  \textbf{Haoxuan Zhang\textsuperscript{1}} \quad
  \textbf{Ruochi Li\textsuperscript{2}} \quad
  \textbf{Yang Zhang\textsuperscript{1,*}} \quad
  \textbf{Zhenni Liang\textsuperscript{1}} \\
  \textbf{Junhua Ding\textsuperscript{1}} \quad
  \textbf{Ting Xiao\textsuperscript{1}} \quad
  \textbf{Haihua Chen\textsuperscript{1,*}} \\
  \textsuperscript{1}University of North Texas, Denton, TX, USA \\
  \textsuperscript{2}North Carolina State University, Raleigh, NC, USA \\
  \texttt{\{haoxuanzhang, zhenniliang\}@my.unt.edu} \\
  \texttt{\{yang.zhang, junhua.ding, ting.xiao, haihua.chen\}@unt.edu} \\
  \texttt{li14@ncsu.edu} \\
  \textsuperscript{*}Corresponding authors
}

\begin{document}
\maketitle
\begin{abstract}

The rapid proliferation of Generative AI necessitates rigorous documentation standards for transparency and governance. However, manual creation of Model and Data Cards is not scalable, while automated approaches lack large-scale, high-fidelity benchmarks for systematic evaluation. We introduce MetaGAI, a comprehensive benchmark comprising 2,541 verified document triplets constructed through semantic triangulation of academic papers, GitHub repositories, and Hugging Face artifacts. Unlike prior single-source datasets, MetaGAI employs a multi-agent framework with specialized Retriever, Generator, and Editor agents, validated through four-dimensional human-in-the-loop assessment, including human evaluation of editor-refined ground truth. We establish a robust evaluation protocol combining automated metrics with validated LLM-as-a-Judge frameworks. Extensive analysis reveals that sparse Mixture-of-Experts architectures achieve superior cost-quality efficiency, while a fundamental trade-off exists between faithfulness and completeness. MetaGAI provides a foundational testbed for benchmarking, training, and analyzing automated Model and Data Card generation methods at scale. Our data and code are available at \url{https://github.com/haoxuan-unt2024/MetaGAI-Benchmark}.
\end{abstract}

\section{Introduction}

The rapid proliferation of Generative AI (GenAI) has fundamentally transformed machine learning deployment, with more than two million models and datasets now hosted on community platforms such as Hugging Face~\cite{wolf2019huggingface, horwitz2025charting}. As these systems transition from research artifacts to critical infrastructure, documentation standards have evolved from foundational Model Cards~\cite{mitchell2019model} and Data Cards~\cite{pushkarna2022data} to comprehensive System Cards~\cite{hurst2024gpt, comanici2025gemini, agarwal2025gpt}, enabling longitudinal model tracking, compliance auditing, and risk assessment in high-stakes domains~\cite{longpre2024large, castano2024analyzing}. Without high-quality documentation, the AI ecosystem lacks the interoperability required to benchmark capabilities or trace data provenance across complex supply chains~\cite{rahman2025hugginggraph}.

However, a significant bottleneck impedes widespread adoption. The GenAI ecosystem is increasingly driven by the ``long tail'' of open-source contributions, comprising thousands of models on community platforms~\cite{wolf2019huggingface, horwitz2025charting}. Unlike well-resourced industry laboratories, developers in this ecosystem often lack the capacity to maintain rigorous documentation. Manual card creation suffers from severe scalability constraints, characterized by pervasive incompleteness, inconsistency, and subjectivity~\cite{yangnavigating, liang2024systematic}, leading to a transparency crisis where many research papers and repositories lack structured metadata necessary for reproducibility~\cite{olmo2025olmo}. Automated documentation generation has emerged as a critical necessity~\cite{liu2024automatic}. To address this challenge, current approaches face significant hurdles: zero-shot methods often hallucinate details when summarizing lengthy documents, while retrieval-augmented strategies struggle to align diverse paper structures with rigid schemas. Progress is stalled by the lack of large-scale, high-quality benchmarks that can objectively measure automated generation accuracy against verified ground truth.

To fill this gap, we introduce \textbf{MetaGAI}, a large-scale benchmark designed to systematically evaluate automated Model and Data Card generation. Unlike prior datasets treating documentation as simple summarization, MetaGAI formulates it as a complex multi-source information generation task, mirroring real-world requirements for verifying scientific claims against implementation details.

\begin{figure}[t]
    \centering
    \includegraphics[width=\linewidth]{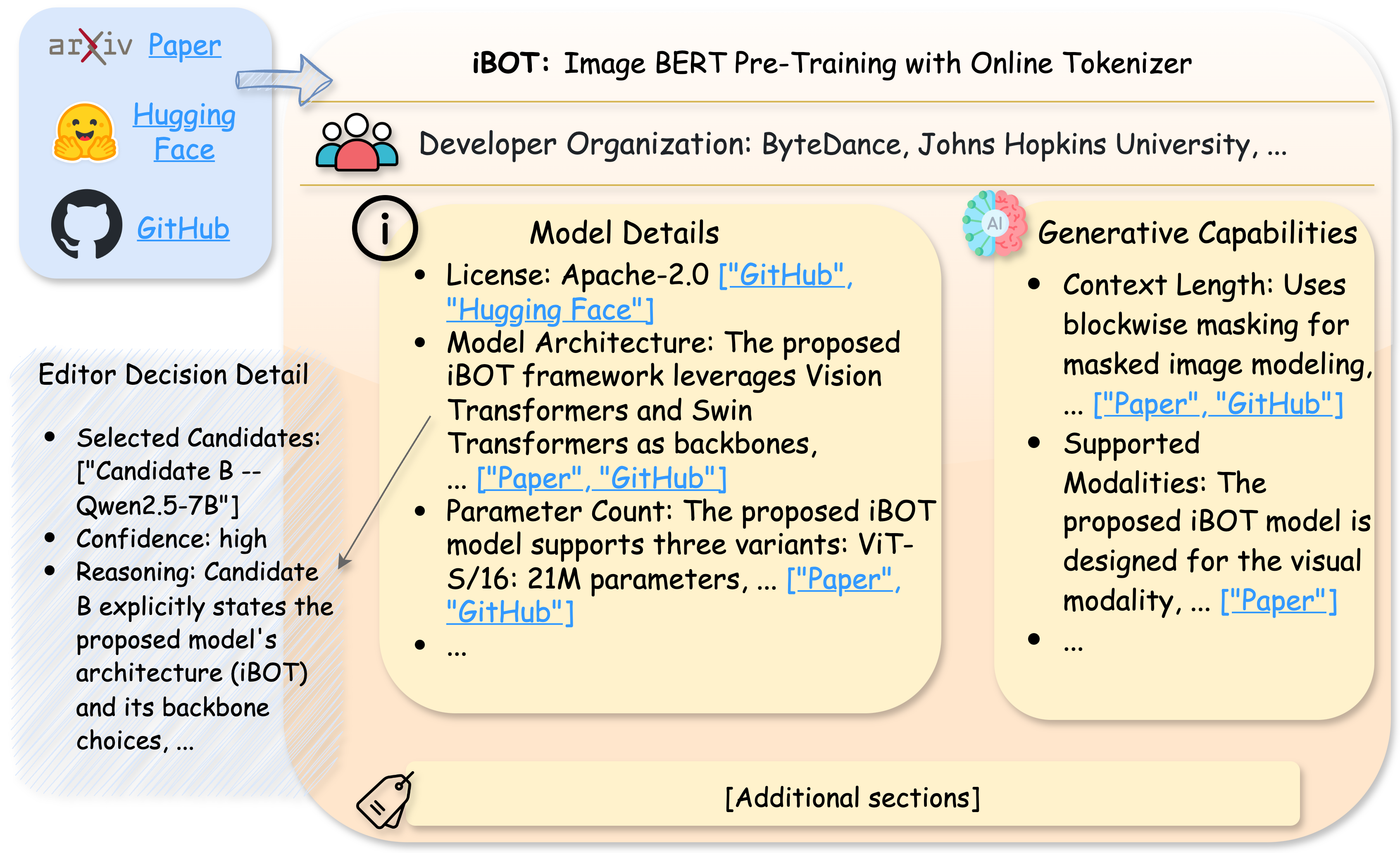}
    \caption{MetaGAI Benchmark Construction Example. Automated GenAI card generation for the iBOT model~\cite{zhouimage} demonstrating Multi-Source Triangulation combining architectural concepts from Papers, hyperparameters from GitHub, and licensing data from Hugging Face, with Editor-Based Synthesis to produce high-fidelity ground truth.}
    \label{fig:intro-demo}
\end{figure}

As illustrated in Figure~\ref{fig:intro-demo}, creating complete documentation cards requires triangulating evidence from heterogeneous sources. Taking the iBOT model~\cite{zhouimage} as an example, architectural concepts are derived from the academic Paper, implementation details are extracted from the GitHub repository, and deployment constraints are verified via Hugging Face. Our construction pipeline utilizes a multi-agent framework comprising specialized Retriever, Generator, and Editor agents to synthesize these signals into verified ground truth. We implement rigorous human-in-the-loop validation across four dimensions: (1) retrieval strategy validation through domain expert annotation, (2) generator divergence analysis validating ensemble diversity, (3) benchmark ground truth quality assessment through hybrid human-LLM evaluation of editor-refined ground truth, and (4) editor architecture selection through pairwise comparisons, establishing a rigorous standard for evaluating automated systems.

In a nutshell, the key contributions of this study are summarized as follows:
\begin{itemize}
    \item We construct MetaGAI, the largest high-quality benchmark for GenAI documentation, comprising 2,541 verified triplets with rigorous human-in-the-loop validation across four dimensions.
    \item We propose a robust evaluation framework combining granular automated metrics with a validated LLM-as-a-Judge protocol.
    \item We provide extensive empirical analysis revealing that sparse Mixture-of-Experts (MoE) architectures offer superior cost-quality efficiency, though a systematic trade-off persists between faithfulness and completeness in generating metadata.
\end{itemize}

\section{Related Work}
\textbf{Model Documentation.}
The lack of standardized documentation for trained machine learning models limits transparency, reproducibility, and responsible use in Natural Language Processing (NLP). Model Cards were introduced to provide structured summaries of trained models, including intended use, evaluation settings, performance, and limitations \citep{mitchell2019model}. Dataset-level documentation was developed in parallel. Data Cards focus on data provenance, collection processes, representativeness, and ethical considerations \citep{pushkarna2022data}. Together, Model Cards and Data Cards form a documentation framework that addresses both model-level and data-level sources of uncertainty. Subsequent work extended the model documentation to interactive and machine-readable formats. Interactive Model Cards were shown to better support user understanding of model behavior \citep{crisan2022interactive}, while linked and semantic representations of documentation were proposed to improve traceability and reuse \citep{donald2023towards}. Empirical analysis of a large number of Model Cards indicates substantial variation in documentation quality, with evaluation and limitation sections often missing or incomplete \citep{liang2024systematic}.

\textbf{Automatic Model and Data Card Generation.}
To reduce the manual effort required for documentation, recent work has explored automatic generation of Model and Data Cards. \citet{liu2024automatic} introduced CARDBENCH, a benchmark of human-written cards, together with CARDGEN, an LLM-based system that generates structured documentation by retrieving information from model repositories and associated papers. Their results show that automatic generation can improve documentation completeness. However, existing benchmarks remain limited in scale and coverage, which constrains systematic evaluation of generation methods. To address this limitation, we introduce MetaGAI, a large-scale benchmark for GenAI Model and Data Card generation.

\begin{figure*}[ht!]
    \centering
    \includegraphics[width=0.9\linewidth]{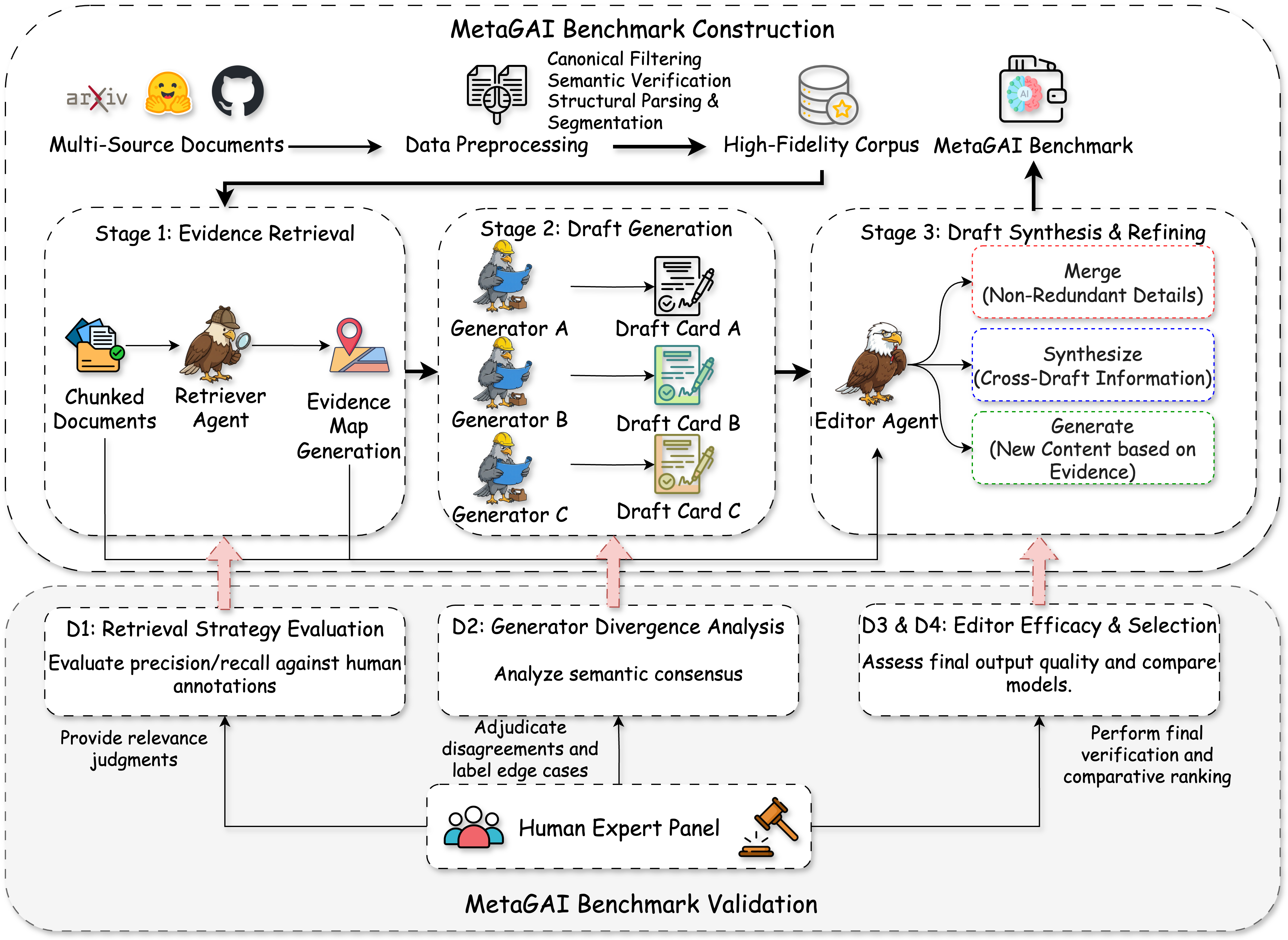}
    \caption{\textbf{MetaGAI Benchmark Construction and Validation Framework.} The pipeline integrates multi-source document preprocessing, a multi-agent generation framework (Evidence Retrieval, Draft Generation, Draft Synthesis and Refining), and a four-dimensional validation protocol (D1-D4) incorporating human expert adjudication.}
    \label{fig:framework}
\end{figure*}

\section{MetaGAI Benchmark Construction}

Figure~\ref{fig:framework} illustrates the MetaGAI benchmark construction and validation pipeline, comprising three sequential stages: (1) evidence retrieval from multi-source documents, (2) multi-generator draft synthesis, and (3) editor-based consolidation and refinement. This architecture extends beyond traditional single-model approaches~\cite{liu2024automatic} by incorporating ensemble generation and rigorous human-in-the-loop validation to ensure high-fidelity card generation for GenAI ecosystems.

\subsection{Task Definition}
\label{sec:task_definition}

We formulate \textbf{MetaGAI} as a structured generation task mapping unstructured scientific papers $\mathcal{P}$ to standardized cards $\mathcal{C}$. The benchmark embodies a deliberate asymmetry between ground truth construction and evaluation input. The ground truth $\mathcal{C}_{GT}$ is derived from a document triplet $(\mathcal{P}, \mathcal{G}, \mathcal{H})$, where $\mathcal{G}$ denotes the GitHub repository and $\mathcal{H}$ the Hugging Face artifact, to capture the implementation details frequently omitted in manuscripts~\cite{mitchell2019model,pushkarna2022data}. The evaluation objective, however, is restricted to paper input $\mathcal{P}$ alone, reflecting the most prevalent real-world deployment scenario and allowing scalable, fair, and difficulty-appropriate evaluation across artifacts.

We adopt and extend the taxonomies of \citet{mitchell2019model} and \citet{pushkarna2022data} with GenAI-specific fields (Appendix~\ref{sec:definition}, Table~\ref{tab:definition_academic}). Each card $\mathcal{C}$ comprises $M$ field-value pairs $\{ (k_i, v_i) \}_{i=1}^M$, where $k_i$ is a schema attribute and $v_i$ its content. The objective is to learn a mapping $f_\theta$ predicting $\hat{\mathcal{C}}$ from paper input alone:

\begin{equation}
\label{eq:paper_to_card}
\hat{\mathcal{C}} = f_\theta(\mathcal{P})
\end{equation}

\noindent minimizing divergence from the multi-source reference $\mathcal{C}_{GT}(\mathcal{P}, \mathcal{G}, \mathcal{H})$, thereby quantifying how much documentation quality can be recovered under paper-only constraints.

\subsection{Data Acquisition and Filtering}
\label{sec:data_acquisition}

\textbf{Corpus Construction.} We construct the benchmark through systematic triangulation of arXiv, GitHub, and Hugging Face metadata to capture implementation details absent from published papers. Unlike previous work that aggregates existing documentation~\cite{liu2024automatic}, our work constructs a new corpus from scratch, with ground truth verified via cross-source semantic consistency. An initial corpus of \textbf{15,727} candidates is canonically filtered to identify unique paper–model–dataset linkages, resulting in \textbf{4,068} entries.

\textbf{Semantic Verification.} To eliminate spurious citations and ensure cross-source alignment, we employed Qwen3-30B-A3B-Instruct~\cite{yang2025qwen3} for automated semantic consistency verification across the three sources (Prompt~\ref{appendix:prompt_correspondence}). To validate this automated approach, two domain experts independently performed a binary correspondence judgment on a pilot sample of 100 triplets (50 model, 50 dataset): each annotator was asked solely whether the Paper, GitHub repository, and Hugging Face artifact referred to the same entity, without generating or evaluating any card content. This objective correspondence task yielded perfect inter-annotator agreement, confirming that source-level entity matching is unambiguous when triplets are well-formed and validating the reliability of the automated filtering procedure. The full filtering pipeline produced a final high-fidelity corpus of \textbf{2,541} verified triplets (Appendix~\ref{appendix:statistics}). Detailed provenance analysis appears in Appendix~\ref{sec:provenance_dynamics}.

\subsection{Benchmark Generation Pipeline}
\label{sec:pipeline}

We developed an automated pipeline synthesizing high-fidelity cards through document preprocessing and a multi-agent framework.

\subsubsection{Pre-processing}
Raw PDF documents were converted to structured Markdown using OLMoCR-2, a state-of-the-art OCR model optimized for academic layouts~\cite{poznanski2025olmocr}. To accommodate context window constraints, the converted papers and README files were segmented into 1024-token chunks with 10\% overlap between adjacent chunks to preserve semantic continuity. Chunk boundaries are always extended to the nearest sentence boundary to avoid splitting semantically coherent passages, ensuring accurate retrieval of implementation details dispersed throughout documents.

\subsubsection{Multi-Agent Framework}
\label{sec:multi_agent}
We employ a multi-agent architecture (Algorithm~\ref{alg:metagai_pipeline}) to maximize factual grounding and minimize hallucinations~\cite{du2023improving}. Where traditional approaches rely on single-model generation following retrieval~\cite{liu2024automatic}, our framework introduces ensemble diversity and editor-based cross-validation. The framework processes full context $\mathcal{P} \cup \mathcal{S}$, where $\mathcal{S} = \{\mathcal{G}, \mathcal{H}\}$ denotes supplementary documentation, through three specialized agents:

\begin{align}
v_i &= \text{Retriever}(\mathcal{P} \cup \mathcal{S}, k_i) \\
\tilde{\mathcal{C}} &= \text{Generator}(\{ (k_i, v_i) \}_{i=1}^M) \\
\hat{\mathcal{C}} &= \text{Editor}(\mathcal{P} \cup \mathcal{S}, \tilde{\mathcal{C}})
\end{align}

\paragraph{Retriever Agent.} This agent aligns unstructured text with schema fields through exhaustive chunk-level classification. Each document segment receives a relevance score (0–4) along with matched sub-fields selected from a predefined ontology (Prompt~\ref{appendix:prompt_retriever}). Comparative analysis (Section~\ref{sec:retriever_val}) established generative models' superiority over discriminative rerankers for complex schema alignment, informing our selection of Qwen3-30B-A3B-Instruct as the retrieval backbone. This generative reasoning approach contrasts with embedding-based similarity matching, enabling more nuanced interpretation of implicit metadata requirements.

\paragraph{Generator Agent.}
To mitigate architectural bias and capture diverse interpretations, we employ ensemble generation~\cite{wangself} with three architecturally distinct LLMs (OLMo-3-7B, Llama-3.1-8B, Qwen2.5-7B). Each model independently synthesizes the draft content, evidence quotations, and confidence scores. A strict ``Subject Focus'' constraint (Prompt~\ref{appendix:prompt_generator}) ensures agents synthesize information about the proposed model or dataset only, excluding baselines and prior work. This multi-model design addresses single-architecture limitations in capturing the full evidence spectrum across complex GenAI systems.

\paragraph{Editor Agent.}
A ``Chief Editor'' consolidates candidate drafts using a larger, cross-family model to avoid self-enhancement bias~\cite{zheng2023judging}. The editor validates semantic alignment between drafts and raw evidence, filters attribution errors, and merges non-redundant details into concise final entries (Prompt~\ref{appendix:prompt_editor}). This explicit cross-validation stage against original sources provides stronger hallucination mitigation than single-pass generation, which is particularly critical for technical specifications where factual precision is essential. Concrete examples appear in Appendix~\ref{sec:case_studies}.

\section{MetaGAI Validation and Analysis}
\label{quality_eva}

To rigorously validate the benchmark construction pipeline, we conduct human-in-the-loop experimentation in four critical dimensions, substantially expanding beyond prior validation efforts~\cite{liu2024automatic} through systematic component-wise evaluation. Our validation framework examines retrieval strategies (\textbf{D1}), quantifies semantic variation between ensemble generators (\textbf{D2}), measures quality improvements from editor-based consolidation (\textbf{D3}), and compares performance between editor architectures (\textbf{D4}).

\subsection{Retrieval Strategy Validation (D1)}
\label{sec:retriever_val}

We compared five retrieval models across two paradigms on a human-verified gold standard of 276 unanimously agreed chunk-field decisions (Appendix~\ref{appendix:retriever_protocol}). As shown in Table~\ref{tab:retriever_val}, discriminative rerankers achieve P@1 $< 0.14$, indicating that semantic similarity alone is insufficient for schema mapping. Qwen3-30B-A3B-Instruct achieves the best performance (P@1: 0.635, F1@5: 0.342), establishing it as the retrieval backbone.

\begin{table}[t]
    \centering
    \resizebox{\linewidth}{!}{%
    \begin{tabular}{l c c c}
        \toprule
        \textbf{Model} & \textbf{P@1} & \textbf{R@5} & \textbf{F1@5} \\
        \midrule
        \multicolumn{4}{l}{\textit{Discriminative Rerankers}} \\
        \quad BGE-Reranker-v2-m3       & 0.135 & 0.114 & 0.101 \\
        \quad Qwen3-Reranker-8B        & 0.115 & 0.137 & 0.091 \\
        \midrule
        \multicolumn{4}{l}{\textit{Generative LLMs}} \\
        \quad Llama-3.1-8B-Instruct    & 0.615 & 0.406 & 0.326 \\
        \quad Qwen2.5-7B-Instruct      & 0.538 & 0.282 & 0.232 \\
        \quad Qwen3-30B-A3B-Instruct   & \textbf{0.635} & \textbf{0.469} & \textbf{0.342} \\
        \bottomrule
    \end{tabular}%
    }
    \caption{Comparative validation of retrieval models (D1). Best results in bold.}
    \label{tab:retriever_val}
\end{table}

\subsection{Generator Divergence Analysis (D2)}
\label{sec:consistency_analysis}

We analyzed semantic consensus within the generator ensemble (OLMo-3-7B, Llama-3.1-8B, Qwen2.5-7B) by computing pairwise BERTScore~\cite{zhangbertscore} similarity across all 2,541 samples. Fields clustering in the high-consensus zone ($\mu > 0.8$), such as \textit{Ethical Considerations} ($\mu = 0.848$), predominantly reflect evidence scarcity rather than genuine agreement (Appendix~\ref{sec:sparsity_analysis}). Conversely, substantial portions of both card types exhibit pronounced divergence ($\sigma > 0.4$), confirming that architecturally distinct models generate complementary details from information-rich contexts and validating our ensemble design. Full mean-variance analysis appears in Appendix~\ref{sec:generator_divergence_full}.

\subsection{Benchmark Ground Truth Quality Assessment}
\label{sec:humaneval}

A benchmark is only as trustworthy as its ground truth: evaluation results are meaningful only if the reference cards are of sufficiently high quality that observed effect sizes exceed the noise floor. To directly address this concern, we conducted a controlled human quality study assessing the final editor-refined cards, which constitute the MetaGAI ground truth, against an unedited generation baseline. We compare three outputs on a stratified random sample of 100 cards (50 models, 50 dataset): (1)~\textbf{Raw Baseline}, the highest-token-count unedited generator draft; (2)~\textbf{Bench\_Mistral}, the output of the Mistral-3-14B-Instruct editor; and (3)~\textbf{Bench\_GPT-OSS}, the output of the GPT-OSS-20B editor. The complete annotation protocol, judge configuration, and inter-annotator agreement are reported in Appendix~\ref{appendix:humaneval_protocol}.

\begin{table}[t!]
\centering
\resizebox{\columnwidth}{!}{%
\begin{tabular}{l ccccc}
\toprule
\textbf{System} & \textbf{Ann-1} & \textbf{Ann-2} & \textbf{Llama} & \textbf{GPT-OSS} & \textbf{Qwen} \\
\midrule
Bench\_GPT-OSS  & \textbf{4.475} & 3.669          & 4.101          & \textbf{3.019} & \textbf{4.303} \\
Bench\_Mistral  & 4.355          & \textbf{3.799} & \textbf{4.634} & 2.897          & 3.990 \\
\midrule
Raw Baseline    & 3.709          & 2.966          & 3.130          & 2.951          & 3.447 \\
\bottomrule
\end{tabular}}
\caption{Mean Likert scores (1--5) for benchmark ground-truth card quality across two domain annotators and three LLM judges on a 100-card stratified sample. \textbf{Bold} denotes the best score per column among editor-refined systems.}
\label{tab:editor_quality}
\end{table}

\paragraph{Editor Efficacy (D3).}
Table~\ref{tab:editor_quality} reports the mean Likert scores (1--5) in two domain annotators and three LLM judges. Editor-refined ground-truth cards achieve mean human scores of $4.41$ (Bench\_GPT-OSS: Ann-1 $4.48$, Ann-2 $3.67$; Bench\_Mistral: Ann-1 $4.36$, Ann-2 $3.80$), substantially above the Raw Baseline ($3.71$ / $2.97$). This $15$--$20\%$ absolute gain on a 5-point scale is well above any noise threshold that would render small effect sizes uninterpretable, confirming that editor consolidation reliably elevates ground-truth quality across all independent evaluators.

\paragraph{Editor Architecture Selection (D4).}
Bench\_GPT-OSS and Bench\_Mistral achieve near-equivalent quality: Bench\_GPT-OSS leads marginally under Ann-1 ($4.48$ vs.\ $4.36$), while Bench\_Mistral scores highest among LLM judges via Llama-3.3-70B-Instruct ($4.63$ vs.\ $4.10$); average annotator scores are indistinguishable ($4.07$ vs.\ $4.08$). Given this parity, we incorporate both architectures and average their outputs as the final benchmark ground truth to ensure architectural neutrality.

\section{Experiments}
\label{sec:experiments}

\subsection{Experimental Setup}

We evaluate all models with zero-shot, generating cards $\hat{\mathcal{C}}$ exclusively from paper text $\mathcal{P}$, consistent with the task definition in Section~\ref{sec:task_definition}. This reflects realistic deployment constraints, as many published artifacts lack repositories or Hugging Face pages entirely~\cite{liang2024systematic,yangnavigating}. To balance statistical rigor with computational feasibility, automated metrics are calculated on the \textbf{full benchmark} ($N=2{,}541$), while LLM-as-a-Judge evaluations use a \textbf{stratified random sample} of 500 entries (250 each).

\subsection{Baselines}

We examine models categorized by access mode and architectural paradigm. Preliminary experiments with models below 14B parameters reveal consistent failures in producing structured outputs; therefore, we restrict evaluation to architectures with 20B+ parameters.

\textbf{Open-Weight Models.}
We evaluate two architectural classes. \textbf{Dense Models} employ standard transformer architectures (20B--32B parameters), including Mistral-Small-3.2-24B-Instruct, Gemma-3-27B-IT~\cite{gemma_2025}, and Qwen3-32B~\cite{yang2025qwen3}. \textbf{MoE Models} utilize Mixture-of-Experts architectures providing high parameter counts with efficient active parameter usage, including GPT-OSS-20B~\cite{agarwal2025gpt}, NVIDIA-Nemotron-3-Nano-30B-A3B~\cite{nvidia_nemotron_nano_v3_2025}, and Qwen3-30B-A3B-Instruct~\cite{yang2025qwen3}. \textbf{Closed-Source Models.} We include proprietary models accessed via API: GPT-5.1 series (Mini/Nano)\footnote{\url{https://openai.com/index/gpt-5-1/}} and Gemini-2.5 series (Flash/Flash-Lite)~\cite{comanici2025gemini}.

\subsection{Evaluation Metrics}

We employ a dual-layered evaluation combining quantitative structural metrics with qualitative expert judgment.

\paragraph{Automated Metrics.}
We measure recall and semantic alignment through three metrics:
\begin{itemize}
    \item \textbf{Completeness:} Quantifies field-level recall. Let $\mathcal{K}(\mathcal{C})$ denote the set of populated keys in card $\mathcal{C}$:
    \begin{equation}
    \text{Completeness} = \frac{|\mathcal{K}(\hat{\mathcal{C}}) \cap \mathcal{K}(\mathcal{C}_{GT})|}{|\mathcal{K}(\mathcal{C}_{GT})|}
    \end{equation}
    \item \textbf{Semantic Similarity:} We report \textbf{ROUGE-L}~\cite{lin2004rouge} for lexical overlap and \textbf{BERTScore (F1)}~\cite{zhangbertscore} for semantic alignment.
\end{itemize}

\paragraph{LLM-as-a-Judge Evaluation.}
To capture nuances beyond n-gram matching, we implement an ensemble framework using GPT-OSS-120B~\cite{agarwal2025gpt}, Llama-3.3-70B-Instruct~\cite{grattafiori2024llama}, and Qwen3-235B-A22B-2507~\cite{yang2025qwen3}. Following the protocol in Appendix~\ref{appendix:eval_metrics} (Prompt~\ref{appendix:prompt_judge}), judges evaluate the generated fields on a 1--5 Likert scale. Scores are averaged among judges to mitigate bias in the single-model (agreement analysis in Appendix~\ref{appendix:judge_agreement}).

\paragraph{Cost Efficiency.}

To assess economic feasibility at scale, we introduce a Cost Index that estimates the inference cost per card generation task. The index normalizes costs across models by standardizing token consumption to 1M input tokens and 0.2M output tokens, enabling direct price comparison independent of model-specific tokenization efficiency. Pricing is derived from standardized rates on OpenRouter\footnote{\url{https://openrouter.ai/}}, facilitating economic comparison between proprietary and open-weight models.

\subsection{Experimental Results}

\begin{table*}[t!]
    \centering
    \setlength{\tabcolsep}{1.2pt}
    \resizebox{\linewidth}{!}{%
    \begin{tabular}{l cc cc cc cc cc cc cc cc cc c}
        \toprule
        \multirow{2.5}{*}{\textbf{Model}} & \multicolumn{2}{c}{\textbf{Faithfulness}} & \multicolumn{2}{c}{\textbf{Relevance}} & \multicolumn{2}{c}{\textbf{Accuracy}} & \multicolumn{2}{c}{\textbf{Consistency}} & \multicolumn{2}{c}{\textbf{Usefulness}} & \multicolumn{2}{c}{\textbf{Qual. Avg}} & \multicolumn{2}{c}{\textbf{Comp.}} & \multicolumn{2}{c}{\textbf{BScore}} & \multicolumn{2}{c}{\textbf{RL}} & \multirow{2.5}{*}{\textbf{Cost}} \\
        \cmidrule(lr){2-3} \cmidrule(lr){4-5} \cmidrule(lr){6-7} \cmidrule(lr){8-9} \cmidrule(lr){10-11} \cmidrule(lr){12-13} \cmidrule(lr){14-15} \cmidrule(lr){16-17} \cmidrule(lr){18-19}
         & \textbf{D} & \textbf{M} & \textbf{D} & \textbf{M} & \textbf{D} & \textbf{M} & \textbf{D} & \textbf{M} & \textbf{D} & \textbf{M} & \textbf{D} & \textbf{M} & \textbf{D} & \textbf{M} & \textbf{D} & \textbf{M} & \textbf{D} & \textbf{M} & \\
        \midrule
        \multicolumn{20}{l}{\textit{\textbf{Dense Models}}} \\
        Mistral-Small-24B & 3.39 & 3.75 & 3.28 & 3.59 & 3.33 & 3.70 & 3.64 & 3.97 & 2.67 & 3.14 & 3.26 & 3.63 & .280 & .327 & \textbf{.193} & \underline{.194} & \textbf{.270} & \underline{.255} & 0.10 \\
        Gemma-3-27B & 3.81 & 3.97 & 3.76 & 3.94 & 3.76 & 3.96 & 4.03 & 4.23 & 3.26 & 3.58 & 3.72 & 3.94 & .464 & \underline{.734} & .151 & .151 & .214 & .203 & \underline{0.07} \\
        Qwen3-32B & 3.40 & 3.72 & 3.29 & 3.58 & 3.32 & 3.69 & 3.66 & 3.94 & 2.65 & 3.10 & 3.26 & 3.60 & .386 & .513 & \underline{.184} & \textbf{.198} & \underline{.267} & \textbf{.265} & 0.13 \\
        \midrule
        \multicolumn{20}{l}{\textit{\textbf{MoE Models}}} \\
        GPT-OSS-20B & 3.18 & 3.49 & 3.09 & 3.36 & 3.11 & 3.45 & 3.45 & 3.74 & 2.49 & 2.87 & 3.06 & 3.38 & .394 & .492 & .146 & .133 & .235 & .220 & \textbf{0.06} \\
        Nemotron-Nano-30B-A3B & 3.45 & 3.80 & 3.41 & 3.76 & 3.39 & 3.77 & 3.71 & 4.03 & 2.88 & 3.35 & 3.37 & 3.74 & \underline{.557} & .644 & .124 & .133 & .199 & .206 & 0.11 \\
        Qwen3-30B-A3B-Instruct & \textbf{4.33} & \textbf{4.50} & \textbf{4.36} & \textbf{4.57} & \textbf{4.31} & \textbf{4.51} & \textbf{4.46} & \textbf{4.67} & \textbf{4.06} & \textbf{4.49} & \textbf{4.30} & \textbf{4.55} & \textbf{.702} & \textbf{.786} & .169 & .174 & .246 & .243 & 0.15 \\
        \midrule
        \multicolumn{20}{l}{\textit{\textbf{Closed-Source Models}}} \\
        GPT-5-Mini & \underline{4.14} & \underline{4.32} & \underline{4.09} & \underline{4.26} & \underline{4.10} & \underline{4.32} & \underline{4.25} & \underline{4.43} & \underline{3.74} & \underline{4.07} & \underline{4.06} & \underline{4.28} & .556 & .216 & .102 & .117 & .185 & .207 & 0.65 \\
        GPT-5-Nano & 3.18 & 3.38 & 3.06 & 3.18 & 3.10 & 3.32 & 3.46 & 3.57 & 2.53 & 2.73 & 3.06 & 3.23 & .383 & .127 & .113 & .149 & .188 & .228 & 0.13 \\
        Gemini-2.5-Flash & 3.88 & 4.08 & 3.80 & 3.97 & 3.82 & 4.06 & 4.04 & 4.19 & 3.27 & 3.69 & 3.76 & 4.00 & .443 & .181 & .160 & .170 & .241 & .246 & 0.80 \\
        Gemini-2.5-Flash-Lite & 3.93 & 4.10 & 3.88 & 4.03 & 3.87 & 4.08 & 4.12 & 4.25 & 3.39 & 3.69 & 3.83 & 4.03 & .511 & .194 & .140 & .144 & .207 & .208 & 0.18 \\
        \bottomrule
    \end{tabular}%
    }
    \caption{\textbf{MetaGAI Benchmark Comprehensive Results.} Performance across Data Card (D) and Model Card (M) generation. 
    Completeness, BERTScore, and ROUGE-L are reported on the full test set (N=2,541). 
    Qual. Avg (Qualitative Average) is evaluated on a sample of 500 entries. \textbf{Bold}/\underline{Underline} indicate best/second-best performance.}
    \label{tab:metagai_full_results}
\end{table*}

\begin{figure*}[t!]
    \centering
    \begin{minipage}{0.48\linewidth}
        \centering
        \includegraphics[width=\linewidth]{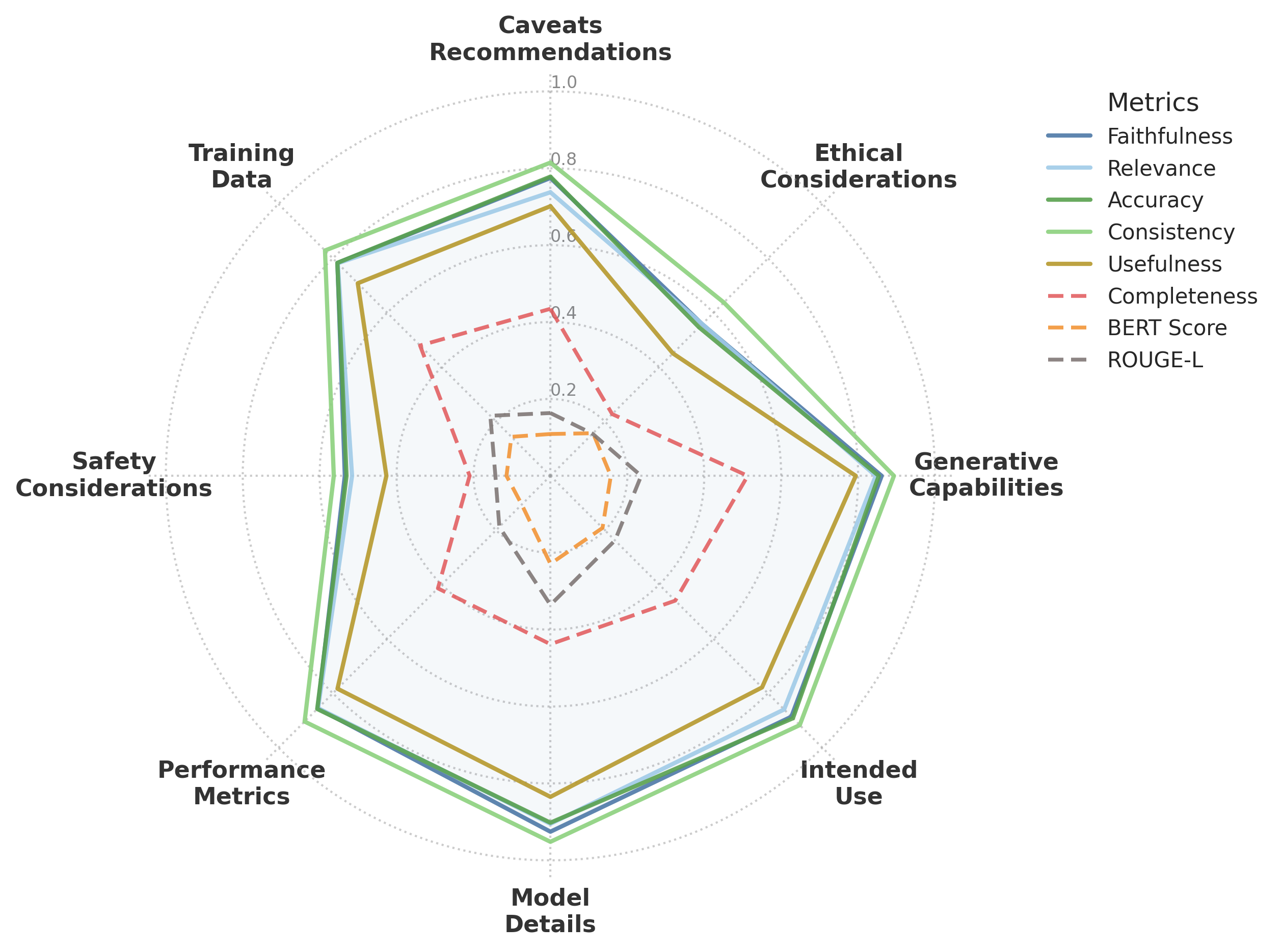}
        \caption*{\small (a) Model Card Fields}
    \end{minipage}
    \hfill
    \begin{minipage}{0.48\linewidth}
        \centering
        \includegraphics[width=\linewidth]{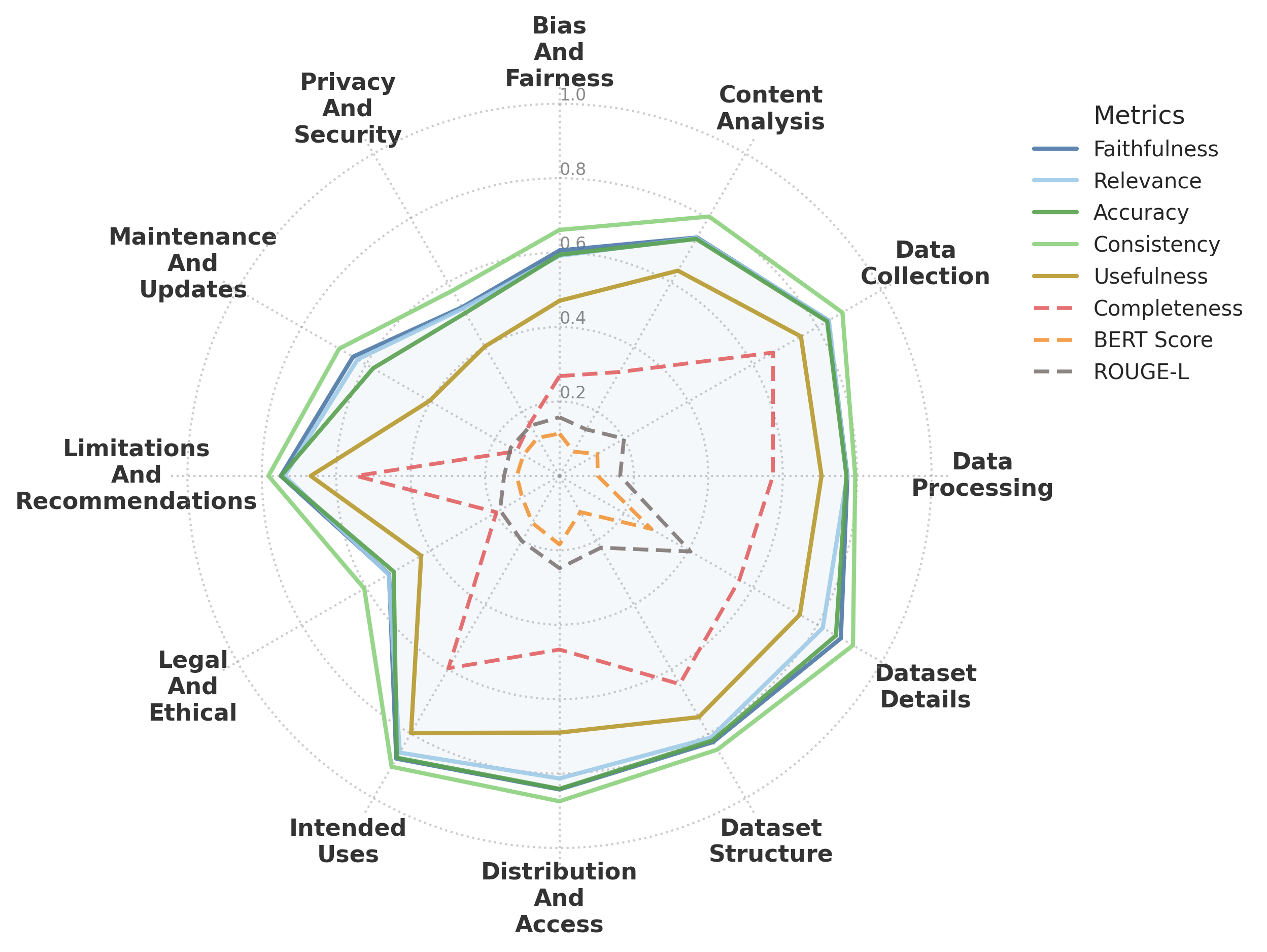}
        \caption*{\small (b) Data Card Fields}
    \end{minipage}
    \caption{\textbf{Field-Level Performance Patterns Averaged Across All Baselines.} Evaluation metrics (colored lines) across card fields (axes). Performance is strong on signal-rich fields (\textit{Model Details}) but degrades on abstract categories (\textit{Ethical Considerations}), revealing systematic generation difficulty when documentation is sparse.}
    \label{fig:radar_charts}
\end{figure*}

We establish a comprehensive evaluation protocol combining automated metrics in 2,541 entries with LLM-as-a-Judge evaluation in 500 samples. Beyond quality assessment, we introduce cost-efficiency analysis, revealing that sparse MoE models achieve optimal cost-quality trade-offs, while traditional lexical metrics inversely correlate with semantic quality. Table~\ref{tab:metagai_full_results} presents comprehensive evaluation results in all baseline models. We analyze performance through architectural efficiency, evaluation metric validity, and cognitive limitations.

\subsubsection{Architectural Efficiency and Economic Viability}
\label{sec:architecture_analysis}

\paragraph{MoE Superiority with Scale Threshold.}
Sparse MoE architectures demonstrate superior parameter efficiency, but only at sufficient scale. Within the Qwen family, MoE-based Qwen3-30B-A3B-Instruct (4.55 Model Card quality) outperforms dense Qwen3-32B (3.60) by 0.95 points despite comparable parameter counts. Given shared pretraining lineages, this gap validates that sparse activation enables more effective information synthesis. However, this advantage requires approximately 30B parameters: the smallest MoE (GPT-OSS-20B) achieves the lowest performance (3.06/3.38), underperforming comparable dense models. This establishes a minimum viable scale threshold for MoE benefits in complex generation tasks.

\paragraph{Open-Weight Cost Dominance.}
Qwen3-30B-A3B-Instruct occupies the Pareto-optimal position on the cost-quality frontier. It delivers state-of-the-art performance (4.55 quality score) at the lowest normalized cost (Cost Index: 0.15), outperforming GPT-5-Mini (4.28 quality, Cost Index: 0.65) by 4.3× in cost efficiency. Gemini-2.5-Flash presents the least favorable value proposition: inferior quality (4.00) at the highest normalized cost (Cost Index: 0.80). Detailed analysis in Appendix~\ref{appendix:cost_efficiency} confirms that optimized open-weight MoE architectures provide economically superior solutions for production-scale card generation.

\subsubsection{Evaluation Metric Validity}
\label{sec:metric_analysis}

\paragraph{Lexical Similarity Metrics Invalidation.}
Traditional content matching metrics exhibit an inverse correlation with semantic quality. ROUGE-L demonstrates this paradox: Mistral-Small-24B achieves the highest ROUGE-L (0.270) through naive verbatim copying, yet produces one of the lowest quality (3.26), while Qwen3-30B-A3B-Instruct yields lower ROUGE-L (0.246) through abstractive synthesis but superior quality (4.30). BERTScore similarly fails as a quality discriminator, compressing all models into a narrow 0.10--0.20 range despite 1.5-point quality differences (3.06--4.55). These metrics penalize good abstractive synthesis that necessarily diverges from the source text.

\paragraph{Completeness-Quality Orthogonality.}
The structural coverage and semantic quality represent independent dimensions. On Model Cards, Gemma-3-27B achieves high Completeness (0.734, second-best) yet scores only 3.94 in quality, substantially below Qwen3-30B-A3B-Instruct (4.55) despite comparable Completeness (0.786). Similarly, on Data Cards, Nemotron-Nano-30B-A3B achieves 0.557 Completeness yet scores only 3.37 in quality, far below Qwen3-30B-A3B-Instruct (4.30) despite the latter's higher Completeness (0.702). This decoupling reveals that high field coverage does not guarantee semantic quality without refinement. Figure~\ref{fig:radar_charts} illustrates field-level performance patterns: averaged across all baselines, models achieve greater than 0.7 Completeness in explicit fields (\textit{Model Details}, \textit{Performance Metrics}) containing abundant signals, but drop below 0.5 in abstract categories (\textit{Ethical Considerations}, \textit{Maintenance}) requiring inference from sparse contexts. Appendix~\ref{sec:sparsity_analysis} shows that ground truth information density explains 31\% of Completeness variance, confirming source sparsity as the primary driver of generation difficulty. These findings validate our multi-metric framework that combines automated structural diagnostics with LLM-as-a-Judge semantic evaluation.

\subsubsection{Systematic Cognitive Limitations}
\label{sec:failure_analysis}

\paragraph{Universal Data Card Difficulty.}
All models show consistent performance degradation on Data Cards versus Model Cards (average gap: 0.2--0.4 points), from weakest (GPT-OSS-20B: 3.06 vs 3.38) to strongest (Qwen3-30B-A3B-Instruct: 4.30 vs 4.55). Data Card schemas demand high-granularity lifecycle documentation (privacy protocols, security measures, maintenance plans), whereas papers treat datasets as ancillary artifacts with sparse experimental descriptions. This systematic scarcity imposes the need to synthesize complete profiles from limited signals, in which current models show consistent degradation in completeness and quality metrics.

\paragraph{Faithfulness-Completeness Trade-off.}
Qwen3-30B-A3B-Instruct achieves near-perfect Faithfulness, avoiding hallucinations, yet exhibits only 0.786 Completeness, systematically omitting 21\% of ground truth fields. This precision-recall imbalance aligns with documented long-context retrieval limitations~\cite{liu2024lost}, where models struggle to access dispersed information. To disentangle whether this gap stems from source sparsity or model inferential capacity, we conducted an ablation study (Appendix~\ref{sec:ablation_provenance}) providing the model with full multi-source context during inference. The results show negligible improvement ($\Delta < 0.02$ among reliable judges), confirming that the completeness bottleneck is attributable to long-context inferential limitations rather than restricted access to evidence. Linguistically, outputs show high abstraction with hedging phrases and cross-references contrasting with the ground truth's concrete specifications. Appendix~\ref{sec:lexical_analysis} quantifies these divergences through Log-Odds Ratio analysis, revealing systematic narrative bias.

\section{Conclusion}
We introduce MetaGAI, a large-scale benchmark comprising 2,541 validated entries constructed through multi-source triangulation and a multi-agent framework. Unlike prior work that aggregates existing documentation, our approach constructs ground truth through rigorous semantic verification and four-dimensional human-in-the-loop validation, addressing incompleteness and inconsistency in human-authored cards. Our experiments reveal that sparse MoE architectures achieve optimal cost-quality performance, while traditional lexical metrics inversely correlate with semantic quality. A fundamental faithfulness-completeness trade-off emerges where models systematically omit fields despite maintaining high factual accuracy. MetaGAI provides a critical infrastructure to advance transparency and reproducibility in GenAI documentation on a production scale.

\clearpage
\section*{Limitations}
\label{sec:limitations}
While MetaGAI provides a strong baseline for automated Model and Data Card generation, it remains limited in scope. Our current framework relies on text-only generation, which overlooks important information embedded in figures, tables, and other non-textual modalities. In addition, MetaGAI treats document generation as isolated paper-level tasks, without modeling the complex dependencies among papers, models, and datasets in the broader GenAI ecosystem. Addressing multimodal content and capturing ecosystem-level relationships through structured or graph-based representations are promising directions for future work. Beyond these technical limitations, automated documentation also introduces potential risks. Errors or omissions in generated cards may propagate misleading signals about model capabilities, data provenance, or licensing conditions, especially when such artifacts are reused at scale. Without careful validation and human oversight, these inaccuracies could undermine transparency efforts or lead to misplaced trust. Addressing multimodal understanding, ecosystem-level modeling, and robust verification mechanisms are important directions for future work.

\section*{Ethics Statement}

We introduce MetaGAI to improve transparency and reproducibility in GenAI through automated documentation. We address the following ethical considerations:

\paragraph{Data Provenance and Licensing.}
Our benchmark triangulates public data from arXiv, GitHub, and Hugging Face, strictly adhering to each platform's terms of use. Source materials are used exclusively for scientific research and card generation. No private information beyond publicly associated author names is collected.

\paragraph{Human Evaluation.}
The validation involved volunteer Ph.D. students in NLP and machine learning. All evaluators were informed about the nature of the task, with a reasonable workload and no exposure to harmful content.

\paragraph{Risks of Automated Documentation.}
We acknowledge the inherent risks of hallucination and omission in LLM-based documentation generation and address them at multiple levels. At the system level, the Generator and Editor Agents produce, for every field, supporting evidence spans, confidence scores, and explicit source attribution, with the Editor verifying attribution against raw source chunks across three independently generated candidates. At the evaluation level, our faithfulness--completeness analysis directly quantifies these failure modes, identifying which fields and models are most susceptible and require prioritized human verification. At the deployment level, generated cards should be treated as structured preliminary drafts subject to human oversight, not as authoritative replacements for expert review.

\paragraph{Environmental Considerations.}
Although the benchmark construction used large-scale LLMs, our cost-quality analysis demonstrates that sparse MoE architectures achieve high-quality generation with substantially lower computational requirements, helping reduce energy consumption in large-scale documentation efforts.

\bibliography{custom}

\begin{thebibliography}{32}
\providecommand{\natexlab}[1]{#1}

\bibitem[{Casta\~{n}o et~al.(2024)Casta\~{n}o, Mart\'{\i}nez-Fern\'{a}ndez, Franch, and Bogner}]{castano2024analyzing}
Joel Casta\~{n}o, Silverio Mart\'{\i}nez-Fern\'{a}ndez, Xavier Franch, and Justus Bogner. 2024.
\newblock \href {https://doi.org/10.1145/3643991.3644898} {Analyzing the evolution and maintenance of ml models on hugging face}.
\newblock In \emph{Proceedings of the 21st International Conference on Mining Software Repositories}, MSR '24, page 607–618, New York, NY, USA. Association for Computing Machinery.

\bibitem[{Chen et~al.(2025)Chen, Xiao, Zhang, Luo, Lian, and Liu}]{chen2024bge}
Jianlv Chen, Shitao Xiao, Peitian Zhang, Kun Luo, Defu Lian, and Zheng Liu. 2025.
\newblock \href {https://arxiv.org/abs/2402.03216} {M3-embedding: Multi-linguality, multi-functionality, multi-granularity text embeddings through self-knowledge distillation}.
\newblock \emph{Preprint}, arXiv:2402.03216.

\bibitem[{Comanici et~al.(2025)Comanici, Bieber, Schaekermann, Pasupat, Sachdeva, Dhillon, Blistein, Ram, Zhang, Rosen, Marris, Petulla, Gaffney, Aharoni, Lintz, Pais, Jacobsson, Szpektor, Jiang, Haridasan, Omran, Saunshi, Bahri, Mishra, Chu, Boyd, Hekman, Parisi, Zhang, Kawintiranon, Bedrax-Weiss, Wang, Xu, Purkiss, Mendlovic, Deutel, Nguyen, Langley, Korn, Rossazza, Ramé, Waghmare, Miller, Byrd, Sheshan, Hadsell, Bhardwaj, Janus, Rissa, Horgan, Abdagic, Belenki, Allingham, Singh, Guidroz, Srinivasan, Schmit, Chiafullo, Elisseeff, Jha, Kolhar, Berrada, Ding, Si, Mallick, Och, Erell, Ni, Latkar, Yang, Sirkovic, Feng, Leland, Hornung, Wu, Blundell, Alvari, Huang, Yip, Deur, Liu, Surita, Duque, Damen, Jia, Guez, Mircea, Sinha, Magni, Stradomski, Marian, Galić, Chen, Husain, Singhal, Grewe, Aubet, Song, Blanco, Rechis, Ho, Munoz, Zheng, Hamrick, Mather, Taitelbaum, Rutherford, Lei, Chen, Shukla, Moreira, Doi, Isik, Shabat, Rogozińska, Kolipaka, Chang, Vušak, Venkatachary, Noghabi, Bharti, Jun, Zaks, Green,
  Challagundla, Wong, Mohammad, Hirsch, Cheng, Naim, Proleev, Vincent, Singh, Krikun, Krishnan, Ghahramani, Atias, Aggarwal, Kirov, Vytiniotis, Koh, Chronopoulou, Dogra, Ion, Tyen, Lee, Weissenberger, Strohman, Balakrishna, Rae, Velic, de~Liedekerke, Elyada, Yuan, Liu, Shani, Kishchenko, Alessio, Li, Song, Kwei, Jankowski, Pappu, Namiki, Ma, Tripuraneni, Cherry, Ikonomidis, Ling, Ji, Westberg, Wright, Yu, Parkinson, Ramaswamy, Connor, Yeganeh, Grover, Kenwright, Litchev, Apps, Tomala, Halim, Castro-Ros, Li, Boral, Sho, Yarom, Malmi, Klinghoffer, Lin, Ansell, S, Zhao, Zuo, Santoro, Cheng, Demmessie, Liu, Brichtova, Culp, Braun, Graur, Ng, Mehta, Phillips, Sundberg, Godbole, Liu, Katariya, Rim, Seyedhosseini, Ammirati, Valfridsson, Malihi, Knight, Toor, Lampe, Ittycheriah, Chiang, Yeung, Fréchette, Rao, Wang, Srivastava, Zhang, Rhodes, Brand, Weesner, Figotin, Gimeno, Fellinger, Marcenac, Leal, Marcus, Cotruta, Cabrera, Luo, Garrette, Axelrod, Baltateanu, Barker, Chen, Toma, Ingram, Riesa, Kulkarni, Zhang,
  Liu, Wang, Polacek, Wu, Hui, Reyes, Su, Barnes, Malhi, Siddiqui, Feng, Damaschin, Pighin, Steiner, Yang, Boppana, Ivanov, Kandoor, Shah, Mujika, Huang, Choquette-Choo, Patel, Yu, Creswell, Jerry, Liu, Barros, Razeghi, Roy, Culliton, Xiong, Pan, Strohmann, Powell, Seal, DeCarlo, Shyam, Katircioglu, Wang, Hardin, Odisho, Broder, Chang, Nair, Shtefan, O'Brien, Agarwal, Potluri, Goyal, Jhindal, Thakur, Stuken, Lyon, Toutanova, Feng, Wu, Horn, Wang, Cullum, Taubman, Shrivastava, Shi, Tomlinson, Patel, Tu, Oflazer, Pongetti, Yang, Taïga, Perot, Pierse, Han, Drori, Iturrate, Chakrabarti, Yeung, Dopson, ting Chen, Kulshreshtha, Guo, Pham, Schuster, Chen, Polozov, Xing, Zhou, Kacham, Kukliansky, Miech, Yaroshenko, Chi, Douglas, Fei, Blondel, Myla, Madmoni, Wu, Keysers, Kjems, Albuquerque, Yu, D'sa, Plantan, Ionescu, Elias, Gupta, Vuyyuru, Alcober, Zhou, Ji, Hartmann, Puttagunta, Song, Amid, Stefanoiu, Lee, Pucciarelli, Wang, Raul, Petrov, Tian, Anklin, Nti, Gomes, Schumacher, Vesom, Panagopoulos, Bousmalis, Andor,
  Jacob, Zhang, Rosgen, Kecman, Tung, Belias, Goodman, Covington, Wieder, Saxena, Davoodi, Huang, Maddineni, Roulet, Campbell-Ajala, Sessa, Xintian, Wu, Lai, Collins, Haig, Sakenas, Xu, Giustina, Shafey, Charoenpanit, Garg, Ainslie, Severson, Arenas, Pathak, Rajayogam, Feng, Bakker, Li, Wichers, Rogers, Geng, Li, Jagerman, Jia, Olmert, Sharon, Mauger, Mariserla, Ma, Mohabey, Kim, Andreev, Pollom, Love, Jain, Agrawal, Schroecker, Fortin, Warmuth, Liu, Leach, Blok, Girirajan, Aharoni, Uria, Sozanschi, Goldberg, Ionita, Ribeiro, Zlocha, Birodkar, Lachgar, Yuan, Choudhury, Ginsberg, Zheng, Dibb, Graves, Lokhande, Rasskin, Muraru, Quick, Tata, Sermanet, Chawla, Karo, Wang, Zhang, Keller, Dragan, Su, Chou, Liu, Tao, Prabhakara, Wilson, Liu, Wang, Evans, Du, Castaño, Prasad, Mahdy, Gerlach, Reid, Kahn, Zait, Pillai, Ulrich, Wang, Wassenberg, Farkash, Yalasangi, Wang, Bauza, Bucher, Liu, Yan, Leung, Sindhwani, Barnes, Singh, Jurin, Chang, Bhumihar, Eiger, Citovsky, Withbroe, Li, Xue, Santo, Stoyanov, Raimond, Zheng,
  Gao, Listík, Kwasiborski, Saputro, Ozturel, Mallya, Majmundar, West, Caron, Wei, Castrejon, Vikram, Ramachandran, Dhawan, Park, Smoot, van~den Driessche, Blau, Malik, Liang, Hirsch, dos Santos, Weinstein, van~den Oord, Lall, FitzGerald, Jiang, Yang, Webster, Elqursh, Pope, Rotival, Raposo, Zhu, Dean, Alabed, Tran, Gupta, Gleicher, Austin, Rosseel, Umekar, Das, Sun, Chen, Misiunas, Zhou, Di, Loo, Newlan, Li, Ramasesh, Xu, Chen, Gandhe, Soricut, Gupta, Hu, El-Sayed, Garcia, Brusilovsky, Chen, Bolt, Huang, Gurney, Zhang, Pritzel, Wilkiewicz, Seybold, Shamanna, Fischer, Dean, Gill, Mcilroy, Bhowmick, Selier, Yang, Cheng, Magay, Tan, Varma, Walder, Kocisky, Nakashima, Natsev, Kwong, Gog, Zhang, Dieleman, Jimma, Ryabtsev, Brahma, Steiner, Du, Žužul, Žanić, Raghavachari, Gierke, Zheng, Petrova, Dauphin, Liu, Kessler, Hand, Duvarney, Kim, Lee, Hussenot, Hui, Smith, Jain, Xia, Tomar, Amiri, Phan, Fuchs, Weyand, Tomasev, Cordell, Liu, Mallinson, Joshi, Crawford, Suggala, Chien, Fernando, Sanchez-Vargas,
  Williams, Crone, Luo, Karpov, Shan, Thurk, Strudel, Voigtlaender, Patil, Dozat, Khodaei, Singla, Ambroszczyk, Wu, Chang, Roark, Hegde, Ding, Filos, Wu, Pinto, Liu, Khanna, Pandey, Mcloughlin, Li, Haves, Zhou, Buchatskaya, Leal, de~Boursac, Akazawa, Anderson, Chen, Somandepalli, Liang, Goenka, Winkler, Grushetsky, Ding, Smith, Ye, Pont-Tuset, Li, Li, Golany, Wegner, Jiang, Barak, Shangguan, Vértes, Wong, Bornschein, Tudor, Bevilacqua, Schaul, Rawat, Zhao, Axiotis, Meng, McLean, Lai, Beattie, Kushman, Liu, Kutzman, Lang, Ye, Netrapalli, Mishra, Khan, Goel, Willoughby, Tian, Zhuang, Chen, Tsai, Kementsietsidis, Khare, Keeling, Xu, Waters, Altché, Popat, Mittal, Saxton, Badawy, Mathieu, Zheng, Zhou, Ranka, Shin, Duan, Salimans, Mihailescu, Shaham, Chang, Assael, Dikkala, Izzard, Cohen-Addad, Graves, Feinberg, Chung, Strouse, Karmon, Sharifzadeh, Ashwood, Pham, Blanton, Vasiloff, Barber, Geller, Zhou, Zubach, Huang, Zhang, Gupta, Young, Proskurnia, Votel, Gabeur, Barcik, Tripathi, Yu, Yan, Changpinyo,
  Pavetić, Coyle, Fujii, Mendez, Zhou, Rajamani, Hechtman, Cao, Juan, Tan, Dalibard, Du, Clay, Yao, Jia, Vijaykumar, Zhou, Bai, Hung, Pecht, Todorov, Khadke, Gupta, Lahoti, Autef, Duddu, Lee-Thorp, Bykovsky, Misiunas, Flennerhag, Thangaraj, McGiffin, Nado, Kunesch, Noever, Hertz, Liang, Stone, Palmer, Daruki, Pramanik, Põder, Kyker, Khan, Sluzhaev, Ritter, Ruderman, Zhou, Nagpal, Vodrahalli, Necula, Barham, Pavlick, Hartford, Shafran, Zhao, Mikuła, Eccles, Shimokawa, Garg, Vilnis, Chen, Shumailov, Lee, Abdelhamed, Xie, Cohen, Hlavnova, Malkin, Sitawarin, Lottes, Coquinot, Yu, Kumar, Zhang, Mahendru, Ahmed, Martens, Chen, Boag, Peng, Devin, Klimovskiy, Phuong, Vainstein, Xie, Ramabhadran, Howard, Yu, Goswami, Cui, Shleifer, Pinto, Yeh, Yang, Javanmardi, Ethier, Lee, Orbay, Kotecha, Bromberg, Shaw, Thornton, Rosenthal, Gu, Thomas, Gemp, Ayyar, Ushio, Selvan, Wee, Liu, Majzoubi, Yu, Abernethy, Liechty, Pan, Nguyen, Qiong, Hu, Perrin, Arora, Pitler, Wang, Shivakumar, Prost, Limonchik, Wang, Gao, Cour, Buch,
  Gui, Ivanova, Neubeck, Chan, Kim, Chen, Goyal, Chung, Liu, Su, Petrushkina, Shen, Joulin, Xu, Lin, Kulizhskaya, Chelba, Vasudevan, Collins, Bashlovkina, Lu, Fritz, Park, Zhou, Su, Tanburn, Sushkov, Rasquinha, Li, Prendki, Li, LV, Sharma, Fitoussi, Huang, Dai, Dao, Burrows, Prior, Qin, Pundak, Sjoesund, Khurshudov, Zhu, Webson, Kemp, Tan, Agrawal, Sargsyan, Cheng, Stephan, Kwiatkowski, Reid, Byravan, Michaely, Heess, Zhou, Goenka, Carpenter, Levskaya, Wang, Roberts, Leblond, Chikkerur, Ginzburg, Chang, Riachi, Chuqiao, Xu, Borsos, Pliskin, Pawar, Lustman, Kirkwood, Anand, Chaudhary, Kalb, Milan, Augenstein, Goldie, Prince, Raman, Sun, Xia, Cohen, Huo, Camp, Ellis, Zilka, Torres, Patel, Arora, Chan, Adler, Ayoub, Liang, Jamil, Jiang, Baumgartner, Sun, Karov, Akulov, Zheng, Cai, Fantacci, Rubin, Acha, Wang, D'Souza, Sathyanarayana, Dai, Rowe, Simanovsky, Goldman, Kuang, Pan, Rosenberg, Rojas-Esponda, Dutta, Zeng, Jurenka, Farquhar, Bansal, Iqbal, Roelofs, Joung, Beak, Ryu, Poplin, Wu, Alayrac, Buthpitiya,
  Ronneberger, Habtegebriel, Li, Cavallaro, Wei, Bensky, Denk, Ganapathy, Stanway, Joshi, Bertolini, Lo, Ma, Charles, Sampemane, Sahni, Chen, Askham, Gaddy, Young, Tan, Eyal, Bražinskas, Zhong, Wu, Epstein, Bailey, Hard, Lee, Goldshtein, Ruiz, Badawi, Lochbrunner, Kearns, Brown, Pardo, Weber, Yang, Jiang, Akin, Fu, Wainwright, Zou, Gaba, Manzagol, Kan, Song, Zainullina, Lin, Ko, Deshmukh, Jindal, Svensson, Tyam, Zhao, Kaeser-Chen, Baird, Moradi, Hall, Guo, Tsang, Liang, Pereira, Ganesh, Korotkov, Adamek, Thiagarajan, Tran, Chen, Tar, Jain, Dasgupta, Bilal, Reitter, Zhao, Vezzani, Gehman, Mehta, Beltrone, Dotiwalla, Guadarrama, Abbas, Karp, Georgiev, Ferng, Brockschmidt, Peng, Hirnschall, Verma, Bi, Xiao, Dabush, Xu, Wallis, Parker, Wang, Xu, Safarli, Tewari, Zhang, Kim, Gesmundo, Thomas, Levi, Chowdhury, Rao, Garst, Conway-Rahman, Ran, McKinney, Xiao, Yu, Agrawal, Stjerngren, Ionescu, Chen, Sharma, Chiu, Liu, Franko, Sanford, Cai, Michel, Ganapathy, Labanowski, Garrett, Vargas, Sun, Gale, Buschmann,
  Desjardins, Ghelani, Jain, Verma, Asawaroengchai, Eisenschlos, Harlalka, Kazawa, Metzler, Howland, Jian, Ades, Shah, Gangwani, Lee, Ring, Hernandez, Reich, Sinha, Sathe, Kovac, Gill, Kannan, D'olimpio, Sevenich, Whang, Kim, Sim, Chen, Zhang, Lall, Matias, Jia, Friesen, Nasso, Thapliyal, Perozzi, Yu, Shekhawat, Huda, Grabowski, Wang, Sreevatsa, Dib, Hassen, Schuh, Milutinovic, Welty, Quinn, Shah, Wang, Barth-Maron, Frye, Axelsson, Zhu, Ma, Giannoumis, Sedghi, Ye, Luan, Aydin, Chandra, Sampathkumar, Huang, Lavrenko, Eleryan, Hong, Hansen, Carthy, Samanta, Ćevid, Wang, Li, Voznesensky, Hoffman, Terzis, Sehwag, Fidel, He, Cai, He, Feng, Nikoltchev, Phatale, Chase, Lawton, Zhang, Ouyang, Tragut, Manshadi, Narayanan, Shen, Gao, Bolukbasi, Roy, Li, Golovin, Panait, Qin, Han, Anthony, Kudugunta, Patraucean, Ray, Chen, Yang, Bhatia, Talluri, Morris, Ražnatović, Brownfield, An, Peng, Kane, Zheng, Duduta, Kessinger, Noraky, Liu, Rong, Veličković, Rush, Goldin, Wei, Garlapati, Pantofaru, Kwon, Ni, Noland, Trapani,
  Beaufays, Roy, Chow, Turker, Cideron, Mei, Clark, Dou, Bošnjak, Leith, Du, Yazdanbakhsh, Nasr, Kwak, Sheth, Kaskasoli, Anand, Lakshminarayanan, Jerome, Bieber, Chu, Senges, Shen, Sridhar, Ndebele, Beyret, Mohamed, Chen, Freitag, Guo, Liu, Roit, Chen, Yan, Stone, Co-Reyes, Cole, Scellato, Azizi, Hashemi, Jin, Iyer, Valentine, György, Ahuja, Diaz, Lee, Clement, Kong, Garmon, Watts, Bhatia, Gupta, Miecnikowski, Vallet, Taly, Loper, Joshi, Atwood, Chick, Collier, Iliopoulos, Trostle, Gunel, Leal-Cavazos, Hrafnkelsson, Guzman, Ju, Forbes, Emond, Chauhan, Caine, Xiao, Zeng, Moufarek, Murphy, Meng, Gupta, Riedel, Das, Lawal, Narayan, Sosea, Swirhun, Friso, Neyshabur, Lu, Girgin, Wunder, Yvinec, Pyne, Carbune, Rijhwani, Guo, Doshi, Briukhov, Bain, Hitron, Wang, Gupta, Chen, Du, Zhang, Shah, Akula, Dylla, Kachra, Kuo, Zou, Wang, Xu, Zhu, Snyder, Menon, Firat, Mordatch, Yuan, Ponomareva, Blevins, Moore, Wang, Chen, Scholz, Dwornik, Lin, Li, Antognini, I, Song, Miller, Kalra, Raveret, Akerlund, Wu, Nystrom, Godbole,
  Liu, DeBalsi, Zhao, Liu, Caciularu, Lax, Khandelwal, Langston, Bailey, Lattanzi, Wang, Kovelamudi, Mondal, Guruganesh, Hua, Roval, Wesołowski, Ingale, Halcrow, Sohn, Angermueller, Raad, Stickgold, Lu, Kosik, Xie, Lillicrap, Huang, Zhang, Paulus, Farabet, Wertheim, Wang, Joshi, ling Ko, Wu, Agrawal, Lin, Sheng, Sung, Breland-King, Butterfield, Gawde, Singh, Zhang, Apte, Shetty, Hutter, Li, Salesky, Lebron, Kanerva, Paganini, Nguyen, Vallu, Peter, Velury, Kao, Hoover, Bortsova, Bishop, Jakobovits, Agostini, Agarwal, Liu, Kwong, Tavakkol, Bica, Greve, GP, Marcus, Hou, Duerig, Moroshko, Lacey, Davis, Amelot, Wang, Kim, Strinopoulos, Wan, Lan, Krishnan, Tang, Humphreys, Bai, Shtacher, Machado, Pang, Burke, Liu, Aravamudhan, Song, Hirst, Singh, Jou, Bai, Piccinno, Fu, Alazard, Meiri, Winter, Chen, Zhang, Heitkaemper, Lambert, Lee, Frömmgen, Rogulenko, Nair, Niemczyk, Bulyenov, Xu, Shemtov, Zadimoghaddam, Toropov, Wirth, Dai, Gollapudi, Zheng, Kurakin, Lee, Bullard, Serrano, Balazevic, Li, Schalkwyk, Murphy,
  Zhang, Sequeira, Datta, Agrawal, Sutton, Attaluri, Chiang, Farhan, Thornton, Lin, Choma, Nguyen, Dasgupta, Robinson, Comşa, Riley, Pillai, Mustafa, Golan, Zandieh, Lespiau, Porter, Ross, Rajayogam, Agarwal, Venugopalan, Shahriari, Yan, Xu, Tobin, Dubov, Shi, Recasens, Kovsharov, Borgeaud, Dery, Vasanth, Gribovskaya, Qiu, Mahdieh, Skut, Nielsen, Zheng, Yu, Bostock, Gupta, Archer, Rawles, Davies, Svyatkovskiy, Tsai, Halpern, Reisswig, Wydrowski, Chang, Puigcerver, Taege, Li, Schnider, Li, Dena, Xu, Telang, Shi, Zen, Kastner, Ko, Subramaniam, Kumar, Blois, Dai, Wieting, Lu, Zeldes, Xie, Hauth, Ţifrea, Li, El-Husseini, Abolafia, Zhou, Ding, Ghalebikesabi, Guía, Maksai, Ágoston Weisz, Arik, Sukhanov, Świetlik, Jia, Yu, Wang, Brand, Bloxwich, Kirmani, Chen, Go, Sprechmann, Kannen, Carin, Sandhu, Edkins, Nooteboom, Gupta, Maggiore, Azizi, Pritch, Yin, Gupta, Tarlow, Smith, Ivanov, Babaeizadeh, Goel, Kambala, Chu, Kastelic, Liu, Soltau, Stone, Agrawal, Kim, Soparkar, Tadepalli, Bunyan, Soh, Kannan, Kim, Chen,
  Halumi, Roy, Wang, Sercinoglu, Gibson, Bhatnagar, Sano, von Dincklage, Ren, Mitrevski, Olšák, She, Doersch, Jilei, Wang, Liu, Tan, Yakar, Warkentin, Ramirez, Lebsack, Dillon, Mathews, Cobley, Wu, Chen, Simon, Nath, Sainath, Bendebury, Julian, Mankalale, Ćurko, Zacchello, Brown, Sodhia, Howard, Caelles, Gupta, Evans, Bulanova, Katzen, Goldenberg, Tsitsulin, Stanton, Schillings, Kovalev, Fry, Shah, Lin, Upadhyay, Li, Radpour, Maggioni, Xiong, Haas, Brennan, Kamath, Savinov, Nagrani, Yacovone, Kappedal, Andriopoulos, Lao, Li, Rozhdestvenskiy, Hashimoto, Audibert, Austin, Rodriguez, Ruoss, Honke, Karkhanis, Xiong, Wei, Huang, Leng, Premachandran, Bileschi, Evangelopoulos, Mensink, Pavagadhi, Teplyashin, Chang, Xue, Tanzer, Goldman, Patel, Li, Wiesner, Zheng, Stewart-Binks, Han, Li, Luo, Lenc, Lučić, Xue, Mullins, Guseynov, Chang, Galatzer-Levy, Zhang, Bingham, Hu, Hartman, Ma, Griffith, Irpan, Radebaugh, Yue, Fan, Ungureanu, Sorokin, Teufel, Li, Anil, Paparas, Wang, Lin, Peng, Shum, Petrovic, Brady,
  Nguyen, Macherey, Li, Singh, Yenugula, Iinuma, Chen, Kopparapu, Stern, Dave, Thekkath, Perot, Kumar, Li, Xiao, Bilotti, Bateni, Noble, Lee, Vázquez-Reina, Salazar, Yang, Wang, Gruzewska, Rao, Raghuram, Xu, Ben-David, Mei, Dalmia, Zhang, Liu, Bansal, Pankov, Schwarcz, Burns, Chan, Sanghai, Liang, Liang, He, Stuart, Narayanan, Zhu, Frank, Fatemi, Sabne, Lang, Bhattacharya, Settle, Wang, McMahan, Tacchetti, Soares, Hadian, Cabi, Chung, Putikhin, Li, Chen, Tarango, Michalewski, Kazemi, Masoom, Sheftel, Shivanna, Vadali, Comanescu, Reid, Moore, Neelakantan, Sander, Herzig, Rosenberg, Dehghani, Choi, Fink, Hayes, Ge, Weng, Ho, Karro, Krishna, Thiet, Skerry-Ryan, Eppens, Andreetto, Sarma, Bonacina, Ayan, Nawhal, Shan, Dusenberry, Thakoor, Gubbi, Nguyen, Tsarfaty, Albanie, Mitrović, Gandhi, Chen, Epasto, Stephanov, Jin, Gehman, Amini, Weber, Behbahani, Xu, Allamanis, Chen, Ott, Sha, Jastrzebski, Qi, Greene, Wu, Toki, Vlasic, Shapiro, Kotikalapudi, Shen, Saeki, Xie, Cassirer, Bharadwaj, Kiyono, Bhojanapalli,
  Rosenfeld, Ritter, Mao, Oliveira, Egyed, Bandemer, Parisotto, Kinoshita, Pluto, Maniatis, Li, Guo, Ghiasi, Tarbouriech, Chatterjee, Jin, Katrina, Xu, Palomaki, Arnold, Sewak, Piccinini, Sharma, Albrecht, Purser-haskell, Vaswani, Chen, Wisniewski, Cao, Aslanides, Phu, Sieb, Agubuzu, Zheng, Sohn, Selvi, Andreassen, Subudhi, Eruvbetine, Woodman, Mery, Krause, Ren, Ma, Luo, Chen, Fan, Griffiths, Schuler, Li, Zhang, Sarr, Luo, Patana, Watson, Naboulsi, Collins, Sidhwani, Hoogeboom, Silver, Caveness, Zhao, Rodriguez, Deines, Bai, Griffin, Tagliasacchi, Xue, Babbula, Pang, Ding, Shen, Peake, Crocker, Raghvendra, Swisher, Han, Singh, Wu, Pchelin, Munkhdalai, Alon, Bacon, Robles, Bulian, Johnson, Powell, Ferreira, Li, Benzing, Velimirović, Soyer, Kong, Tony, Nguyên, Yang, Liu, van Amersfoort, Gillick, Sun, Rauschmayr, Zhang, Zhan, Zhou, Frolov, Yang, Vnukov, Rouillard, Li, Mandhane, Fallen, Venkataraman, Hu, Brennan, Lee, Chang, Sundermeyer, Pan, Ke, Tong, Fabrikant, Bono, Gu, Foley, Mao, Delakis, Bhaswar,
  Frostig, Li, Zipori, Hope, Kozlova, Mishra, Djolonga, Schiff, Merey, Briakou, Morgan, Wan, Hassidim, Skerry-Ryan, Sengupta, Jasarevic, Kallakuri, Kunkle, Brennan, Lieber, Mansoor, Walker, Zhang, Xie, Žužić, Chukwuka, Druinsky, Cho, Yao, Naeem, Butt, Kim, Jia, Jordan, Lelkes, Kurzeja, Wang, Zhao, Over, Chakladar, Prasetya, Jha, Ganapathy, Cong, Shroff, Saroufim, Miryoosefi, Hammad, Nasir, Xi, Gao, Maeng, Hora, Cheng, Haghani, Lewenberg, Lu, Matysiak, Raisinghani, Wang, Baugher, Sukthankar, Giang, Schultz, Fiedel, Chen, Lee, Dey, Zheng, Paul, Smith, Ly, Wang, Bansal, Perz, Ricco, Blank, Keshava, Sharma, Chow, Lad, Jalan, Osindero, Swanson, Scott, Ilić, Li, Jonnalagadda, Soudagar, Xiong, Batsaikhan, Jarrett, Kumar, Shah, Lawlor, Waters, Graham, May, Ramos, Lefdal, Cankara, Cano, O'Donoghue, Borovik, Liu, Grimstad, Alnahlawi, Tsihlas, Hudson, Grigorev, Jia, Huang, Igwe, Lebedev, Tang, Krivokon, Garcia, Tan, Jia, Stys, Vashishth, Liang, Venkatraman, Gu, Kementsietsidis, Zhu, Jung, Bai, Hosseini, Ahmed,
  Gupta, Yuan, Ashraf, Nigam, Vasudevan, Awasthi, Gilady, Mariet, Eskander, Li, Hu, Garrido, Schlattner, Zhang, Saxena, Dević, Muralidharan, Murthy, Zhou, Choi, Wongpanich, Wang, Shah, Xu, Huang, Spencer, Chen, Cohan, Wang, Tompson, Wu, Haroun, Li, Huergo, Yang, Yin, Wendt, Bendersky, Chaabouni, Snaider, Ferret, Jindal, Thompson, Xue, Bishop, Phal, Sharma, Sung, Radhakrishnan, Shomrat, Ingle, Vij, Gilmer, Istin, Sobell, Lu, Nottage, Sadigh, Willcock, Zhang, Xu, Brown, Lee, Wang, Zhu, Tay, Kim, Gutierrez, Sharma, Xian, Seo, Cui, Pochernina, Baetu, Jastrzębski, Ly, Elhawaty, Suh, Sezener, Wang, Yuen, Tucker, Cai, Yang, Wang, Muzio, Qian, Yoo, Lockhart, McKee, Guo, Mehrotra, Mendonça, Mehta, Ben, Tekur, Mu, Zhu, Krakovna, Lee, Maschinot, Cevey, Choe, Bai, Srinivasan, Gasaway, Young, Siegler, Holtmann-Rice, Piratla, Baumli, Yogev, Hofer, van Hasselt, Grant, Chervonyi, Silver, Hogue, Agarwal, Wang, Singh, Flynn, Lipschultz, David, Bellot, Yang, Le, Graziano, Olszewska, Hui, Maurya, Parotsidis, Chen, Oguntebi,
  Kelley, Baddepudi, Mauerer, Shaw, Siegman, Yang, Shetty, Roy, Song, Stokowiec, Burnell, Savant, Busa-Fekete, Miao, Ghosh, MacDermed, Lippe, Dektiarev, Behrman, Mentzer, Nguyen, Wei, Verma, Knutsen, Dasari, Yan, Mitrichev, Wang, Shejwalkar, Austin, Sunkara, Potti, Virin, Wright, Liu, Riva, Pot, Kochanski, Le, Balasubramaniam, Dhar, Liao, Bloniarz, Shukla, Cole, Lee, Zhang, Kafle, Vashishtha, Mahmoudieh, Chen, Hoffmann, Srinivasan, Lago, Shalom, Wang, Elabd, Sharma, Oh, Kothawade, Le, Monteiro, Yang, Alarakyia, Geirhos, Mincu, Garnes, Kobayashi, Mariooryad, Krasowiak, Zhixin, Lai, Mourad, Wang, Bu, Aharoni, Chen, Goyal, Zubov, Bapna, Dabir, Kothari, Lamerigts, Cao, Shar, Yew, Kulkarni, Mahaarachchi, Joshi, Zhu, Lichtarge, Zhou, Muckenhirn, Selo, Vinyals, Chen, Brohan, Mehta, Cogan, Wang, Geri, Ko, Chen, Viola, Shivam, Wang, Elish, Popa, Pereira, Liu, Koster, Kim, Zhang, Ebrahimi, Talukdar, Zheng, Poklukar, Mikhalap, Johnson, Vijayakumar, Omernick, Dibb, Dubey, Hu, Suman, Aggarwal, Kornakov, Xia, Lowe,
  Kolganov, Xiao, Nikolaev, Hemingray, Li, Iljazi, Rybiński, Sandhu, Lu, Luong, Jenatton, Govindaraj, Hui, Li, Dulac-Arnold, Park, Wang, Modi, Pouget-Abadie, Greller, Gupta, Berry, Ramachandran, Xie, McCafferty, Wang, Gupta, Lim, Bratanič, Brock, Akolzin, Sproch, Karliner, Kim, Goedeckemeyer, Shazeer, Schmid, Calandriello, Bhatia, Choromanski, Montgomery, Dua, Ramalho, King, Gao, Nguyen, Lindner, Pitta, Johnson, Salama, Ardila, Han, Farnese, Odoom, Wang, Ding, Rink, Smith, Lehri, Cohen, Vats, He, Gopavarapu, Paszke, Patel, Gansbeke, Loher, Castro, Voitovich, von Glehn, George, Niklaus, Eaton-Rosen, Rakićević, Jue, Perel, Zhang, Bahat, Pouget, Xing, Huot, Shenoy, Bos, Coriou, Richter, Noy, Wang, Ontanon, Qin, Makarchuk, Hassabis, Li, Sharma, Venkatesan, Kemaev, Daniel, Huang, Shah, Ponce, Warren, Chen, Faruqui, Wu, Andačić, Payrits, McDuff, Hume, Cao, Tessler, Wang, Wang, Rendulic, Agustsson, Johnson, Lando, Howard, Padmanabhan, Daswani, Banino, Kilgore, Heek, Ji, Caceres, Li, Kassner, Vlaskin, Liu,
  Grills, Hou, Sukkerd, Cheon, Shetty, Markeeva, Stanczyk, Iyer, Gong, Gao, Gopalakrishnan, Blyth, Reynolds, Bhoopchand, Bilenko, Gharibian, Zayats, Faust, Singh, Ma, Jiao, Vijayanarasimhan, Aroyo, Yadav, Chakera, Kakarla, Meshram, Gregor, Botea, Senter, Jia, Kovacs, Sharma, Baur, Kang, He, Zhuo, Kostelac, Laish, Peng, O'Bryan, Kasenberg, Rao, Leurent, Zhang, Stevens, Salazar, Zhang, Lobov, Walker, Porter, Redshaw, Ke, Rao, Lee, Lam, Moffitt, Kim, Qiao, Koo, Dadashi, Song, Sundararajan, Xu, Kawamoto, Zhong, Barbu, Reddy, Verzetti, Li, Papamakarios, Klimczak-Plucińska, Cassin, Kavukcuoglu, Swavely, Vaucher, Zhao, Hemsley, Tschannen, Ge, Menghani, Yu, Ha, He, Wu, Song, Sterneck, Zinke, Calian, Marsden, Ruiz, Hessel, Gueta, Lee, Farris, Gupta, Li, Saleh, Misra, Xiao, Mendolicchio, Buttimore, Krayvanova, Nayakanti, Wiethoff, Pande, Mirhoseini, Lao, Liu, Hua, Chen, Malkov, Kalashnikov, Gupta, Audhkhasi, Zhai, Kopalle, Jain, Ofek, Meyer, Baatarsukh, Strejček, Qian, Freedman, Figueira, Sokolik, Bachem, Lin,
  Kharrat, Hidey, Xu, Duan, Li, Ersoy, Everett, Cen, Santamaria-Fernandez, Taubenfeld, Mackinnon, Deng, Zablotskaia, Viswanadha, Goel, Yates, Deng, Choy, Chen, Sinha, Mossin, Wang, Szlam, Hao, Rubenstein, Toksoz-Exley, Aperghis, Zhong, Ahn, Isard, Lacombe, Luisier, Anastasiou, Kalley, Prabhu, Dunleavy, Bijwadia, Mao-Jones, Chen, Pasumarthi, Wood, Dostmohamed, Hurley, Simsa, Parrish, Pajarskas, Harvey, Skopek, Kochinski, Rey, Rieser, Zhou, Lee, Acharya, Li, Jiang, Zhang, Gipson, Mahintorabi, Gelmi, Khajehnouri, Yeh, Lee, Matthey, Baker, Pham, Fu, Pak, Gupta, Vasconcelos, Sadovsky, Walker, Hsiao, Zochbauer, Marzoca, Velan, Zeng, Baechler, Driess, Jain, Huang, Tao, Maggs, Levine, Schneider, Gemzer, Petit, Han, Fisher, Zelle, Biles, Ie, Fadeeva, Liu, Franco, Collister, Zhang, Wang, Zhao, Kieliger, Shuster, Zhu, Gong, Chan, Sun, Basu, Zimmermann, Hayes, Bapna, Snoek, Yang, Datta, Abdallah, Kilgour, Li, Mah, Jun, Rivière, Karmarkar, Spalink, Huang, Gonzalez, Tran, Nowak, Palowitch, Chadwick, Talius, Mehta, Sellam,
  Fränken, Nicosia, He, Kini, Amos, Basu, Jobe, Shaw, Xu, Evans, Ikeda, Yan, Jin, Wang, Yadav, Labzovsky, Sampath, Ma, Schumann, Siddhant, Shah, Youssef, Agarwal, Dabney, Tonioni, Ambar, Li, Guyon, Li, Soergel, Fang, Karadzhov, Udrescu, Trinh, Raunak, Noury, Guo, Gupta, Finkelstein, Petek, Liang, Billock, Sun, Wood, Song, Yu, Matejovicova, Cohen, Andra, D'Ambrosio, Deng, Nallatamby, Songhori, Dangovski, Lampinen, Botadra, Hillier, Cao, Baddi, Kuncoro, Yoshino, Bhagatwala, Ranzato, Schaeffer, Liu, Ye, Sarvana, Nham, Kuang, Gao, Baek, Mittal, Wahid, Gergely, Ni, Feldman, Muir, Lamblin, Macherey, Dyer, Kilpatrick, Campos, Bhutani, Fort, Ahmad, Severyn, Chatziprimou, Ferludin, Dimarco, Kusupati, Heyward, Bahir, Villela, Millican, Marcus, Bahargam, Unlu, Roth, Wei, Gopal, Ghoshal, Lee, Lin, Lees, Lee, Hosseini, Fan, Neel, Wu, Altun, Cai, Piqueras, Woodward, Bissacco, Haykal, Bordbar, Sundaram, Hodkinson, Toyama, Polovets, Myers, Sinha, Levinboim, Krishnakumar, Chhaparia, Sholokhova, Gundavarapu, Jawahar, Qureshi,
  Hu, Momchev, Rahtz, Wu, S, Dhamdhere, Guo, Gupta, Eslami, Schain, Blokzijl, Welling, Orr, Bolelli, Perez-Nieves, Sirotenko, Prasad, Kar, Pigem, Terzi, Weisz, Ghosh, Mavalankar, Madeka, Daugaard, Adam, Shah, Berman, Tran, Baker, Andrejczuk, Chole, Raboshchuk, Mirzazadeh, Kagohara, Wu, Schallhart, Orlando, Wang, Rrustemi, Xiong, Liu, Vezer, Ramsden, yiin Chang, Mudgal, Li, Vieillard, Hoshen, Ahmad, Slone, Hua, Potikha, Rossini, Stritar, Prakash, Wang, Dong, Nazari, Nehoran, Tekelioglu, Li, Badola, Funkhouser, Li, Yerram, Ganeshan, Formoso, Langner, Shi, Li, Yamamori, Panda, Saade, Scarpati, Breaux, Carey, Zhou, Hsieh, Bridgers, Butryna, Gupta, Tulsyan, Woo, Eltyshev, Grathwohl, Parks, Benjamin, Panigrahy, Dodhia, Freitas, Sauer, Song, Alet, Tolins, Paduraru, Zhou, Albert, Zhang, Shu, Bansal, Nguyen, Globerson, Xiao, Manyika, Hennigan, Rong, Matak, Bakalov, Sharma, Sinopalnikov, Pierson, Roller, Brown, Gao, Fukuzawa, Ghafouri, Vassigh, Barr, Wang, Korsun, Jayaram, Ren, Zaman, Khan, Lunts, Deutsch, Uthus, Katz,
  Samsikova, Khalifa, Sethi, Sun, Tang, Alon, Luo, Yu, Nayyar, Petrini, Truong, Hellendoorn, Chinaev, Alberti, Wang, Hu, Mirrokni, Balashankar, Aharon, Mehta, Iscen, Kready, Manning, Mohananey, Chen, Tripathi, Wu, Petrovski, Hwang, Baeuml, Chandrakaladharan, Liu, Coaguila, Chen, Ma, Tafti, Tatineni, Spitz, Ye, Vicol, Rosca, Puigdomènech, Yahav, Ghemawat, Lin, Kirk, Nabulsi, Brin, Bohnet, Caluwaerts, Veerubhotla, Zheng, Dai, Petrov, Xu, Mehran, Xu, Zintgraf, Choi, Hombaiah, Thoppilan, Reddi, Lew, Li, Webster, Sawhney, Lamprou, Shakeri, Lunayach, Chen, Bagri, Salcianu, Chen, Donchev, Magister, Nørly, Rodrigues, Izo, Noga, Zou, Köppe, Zhou, Lee, Long, Eisenbud, Chen, Schenck, To, Zhong, Taropa, Truong, Levy, Martins, Zhang, Semturs, Zhang, Yakubovich, Moreno, McConnaughey, Lu, Redmond, Weerts, Bitton, Refice, Lacasse, Conmy, Tallec, Odell, Forbes-Pollard, Socala, Hoech, Kohli, Walton, Wang, Sazanovich, Zhu, Kapishnikov, Galt, Denton, Murdoch, Sikora, Mohamed, Wei, First, McConnell, Cobo, Qin, Avrahami, Balle,
  Watanabe, Louis, Kraft, Ariafar, Gu, Rives, Yoon, Rusu, Cobon-Kerr, Hahn, Luo, Yuvein, Zhu, Ahuja, Benenson, Kaufman, Yu, Hightower, Zhang, Ni, Hendricks, Wang, Yona, Jain, Barrio, Bhupatiraju, Velusamy, Dafoe, Riedel, Thomas, Yuan, Bellaiche, Panthaplackel, Kloboves, Jauhari, Akbulut, Davchev, Gladchenko, Madras, Chuklin, Hill, Yuan, Madhavan, Leonhard, Scandinaro, Chen, Niu, Douillard, Damoc, Onoe, Pedregosa, Bertsch, Leichner, Pagadora, Malmaud, Ponda, Twigg, Duzhyi, Shen, Wang, Garg, Chen, Evci, Lee, Liu, Kojima, Yamaguchi, Rajendran, Piergiovanni, Rajendran, Fornoni, Ibagon, Ragan, Khan, Blitzer, Bunner, Sun, Kosakai, Lundberg, Elue, Guu, Park, Park, Narayanaswamy, Wu, Mudigonda, Cohn, Mu, Kumar, Graesser, Zhang, Killam, Zhuang, Giménez, Jishi, Ley-Wild, Zhai, Osawa, Cedillo, Liu, Upadhyay, Sieniek, Sharma, Paine, Angelova, Addepalli, Parada, Majumder, Lamp, Kumar, Deng, Myaskovsky, Sabolić, Dudek, York, de~Chaumont~Quitry, Nie, Cattle, Gunjan, Piot, Khawaja, Bang, Wang, Khodadadeh, R, Rawlani,
  Powell, Lee, Griesser, Oh, Magalhaes, Li, Tokumine, Vogel, Hsu, BC, Jindal, Cohen, Yang, Yuan, de~Cesare, Bruguier, Xu, Roy, Jacovi, Belov, Arya, Meadowlark, Cohen-Ganor, Ye, Morris-Suzuki, Banzal, Song, Ponnuramu, Zhang, Scrivener, Zaiem, Rochman, Han, Ghazi, Lee, Drath, Suo, Girgis, Shenoy, Nguyen, Eck, Gupta, Yan, Carreira, Gulati, Sang, Mirylenka, Cooney, Chou, Ling, Fan, Coleman, Tubone, Kumar, Baldridge, Hernandez-Campos, Lazaridou, Besley, Yona, Bulut, Wellens, Pierigiovanni, George, Green, Han, Tao, Clark, You, Abdolmaleki, Fu, Chen, Chaugule, Chandorkar, Rahman, Thompson, Koanantakool, Bernico, Ren, Vlasov, Vassilvitskii, Kula, Liang, Kim, Huang, Ye, Lepikhin, and Helmholz}]{comanici2025gemini}
Gheorghe Comanici, Eric Bieber, Mike Schaekermann, Ice Pasupat, Noveen Sachdeva, Inderjit Dhillon, Marcel Blistein, Ori Ram, Dan Zhang, et~al. 2025.
\newblock \href {https://arxiv.org/abs/2507.06261} {Gemini 2.5: Pushing the frontier with advanced reasoning, multimodality, long context, and next generation agentic capabilities}.
\newblock \emph{Preprint}, arXiv:2507.06261.

\bibitem[{Crisan et~al.(2022)Crisan, Drouhard, Vig, and Rajani}]{crisan2022interactive}
Anamaria Crisan, Margaret Drouhard, Jesse Vig, and Nazneen Rajani. 2022.
\newblock \href {https://doi.org/10.1145/3531146.3533108} {Interactive model cards: A human-centered approach to model documentation}.
\newblock In \emph{Proceedings of the 2022 ACM Conference on Fairness, Accountability, and Transparency}, FAccT '22, page 427–439, New York, NY, USA. Association for Computing Machinery.

\bibitem[{Donald et~al.(2023)Donald, Galanopoulos, Curry, Mu\~{n}oz, Ullah, Waskow, Dabrowski, and Kalra}]{donald2023towards}
Andy Donald, Apostolos Galanopoulos, Edward Curry, Emir Mu\~{n}oz, Ihsan Ullah, M.~A. Waskow, Maciej Dabrowski, and Manan Kalra. 2023.
\newblock \href {https://doi.org/10.1145/3543873.3587659} {Towards a semantic approach for linked dataspace, model and data cards}.
\newblock In \emph{Companion Proceedings of the ACM Web Conference 2023}, WWW '23 Companion, page 1468–1473, New York, NY, USA. Association for Computing Machinery.

\bibitem[{Du et~al.(2024)Du, Li, Torralba, Tenenbaum, and Mordatch}]{du2023improving}
Yilun Du, Shuang Li, Antonio Torralba, Joshua~B. Tenenbaum, and Igor Mordatch. 2024.
\newblock \href {https://openreview.net/forum?id=zj7YuTE4t8} {Improving factuality and reasoning in language models through multiagent debate}.
\newblock In \emph{Forty-first International Conference on Machine Learning}.

\bibitem[{Grattafiori et~al.(2024)Grattafiori, Dubey, Jauhri, Pandey, Kadian, Al-Dahle, Letman, Mathur, Schelten, Vaughan, Yang, Fan, Goyal, Hartshorn, Yang, Mitra, Sravankumar, Korenev, Hinsvark, Rao, Zhang, Rodriguez, Gregerson, Spataru, Roziere, Biron, Tang, Chern, Caucheteux, Nayak, Bi, Marra, McConnell, Keller, Touret, Wu, Wong, Ferrer, Nikolaidis, Allonsius, Song, Pintz, Livshits, Wyatt, Esiobu, Choudhary, Mahajan, Garcia-Olano, Perino, Hupkes, Lakomkin, AlBadawy, Lobanova, Dinan, Smith, Radenovic, Guzmán, Zhang, Synnaeve, Lee, Anderson, Thattai, Nail, Mialon, Pang, Cucurell, Nguyen, Korevaar, Xu, Touvron, Zarov, Ibarra, Kloumann, Misra, Evtimov, Zhang, Copet, Lee, Geffert, Vranes, Park, Mahadeokar, Shah, van~der Linde, Billock, Hong, Lee, Fu, Chi, Huang, Liu, Wang, Yu, Bitton, Spisak, Park, Rocca, Johnstun, Saxe, Jia, Alwala, Prasad, Upasani, Plawiak, Li, Heafield, Stone, El-Arini, Iyer, Malik, Chiu, Bhalla, Lakhotia, Rantala-Yeary, van~der Maaten, Chen, Tan, Jenkins, Martin, Madaan, Malo, Blecher,
  Landzaat, de~Oliveira, Muzzi, Pasupuleti, Singh, Paluri, Kardas, Tsimpoukelli, Oldham, Rita, Pavlova, Kambadur, Lewis, Si, Singh, Hassan, Goyal, Torabi, Bashlykov, Bogoychev, Chatterji, Zhang, Duchenne, Çelebi, Alrassy, Zhang, Li, Vasic, Weng, Bhargava, Dubal, Krishnan, Koura, Xu, He, Dong, Srinivasan, Ganapathy, Calderer, Cabral, Stojnic, Raileanu, Maheswari, Girdhar, Patel, Sauvestre, Polidoro, Sumbaly, Taylor, Silva, Hou, Wang, Hosseini, Chennabasappa, Singh, Bell, Kim, Edunov, Nie, Narang, Raparthy, Shen, Wan, Bhosale, Zhang, Vandenhende, Batra, Whitman, Sootla, Collot, Gururangan, Borodinsky, Herman, Fowler, Sheasha, Georgiou, Scialom, Speckbacher, Mihaylov, Xiao, Karn, Goswami, Gupta, Ramanathan, Kerkez, Gonguet, Do, Vogeti, Albiero, Petrovic, Chu, Xiong, Fu, Meers, Martinet, Wang, Wang, Tan, Xia, Xie, Jia, Wang, Goldschlag, Gaur, Babaei, Wen, Song, Zhang, Li, Mao, Coudert, Yan, Chen, Papakipos, Singh, Srivastava, Jain, Kelsey, Shajnfeld, Gangidi, Victoria, Goldstand, Menon, Sharma, Boesenberg,
  Baevski, Feinstein, Kallet, Sangani, Teo, Yunus, Lupu, Alvarado, Caples, Gu, Ho, Poulton, Ryan, Ramchandani, Dong, Franco, Goyal, Saraf, Chowdhury, Gabriel, Bharambe, Eisenman, Yazdan, James, Maurer, Leonhardi, Huang, Loyd, Paola, Paranjape, Liu, Wu, Ni, Hancock, Wasti, Spence, Stojkovic, Gamido, Montalvo, Parker, Burton, Mejia, Liu, Wang, Kim, Zhou, Hu, Chu, Cai, Tindal, Feichtenhofer, Gao, Civin, Beaty, Kreymer, Li, Adkins, Xu, Testuggine, David, Parikh, Liskovich, Foss, Wang, Le, Holland, Dowling, Jamil, Montgomery, Presani, Hahn, Wood, Le, Brinkman, Arcaute, Dunbar, Smothers, Sun, Kreuk, Tian, Kokkinos, Ozgenel, Caggioni, Kanayet, Seide, Florez, Schwarz, Badeer, Swee, Halpern, Herman, Sizov, Guangyi, Zhang, Lakshminarayanan, Inan, Shojanazeri, Zou, Wang, Zha, Habeeb, Rudolph, Suk, Aspegren, Goldman, Zhan, Damlaj, Molybog, Tufanov, Leontiadis, Veliche, Gat, Weissman, Geboski, Kohli, Lam, Asher, Gaya, Marcus, Tang, Chan, Zhen, Reizenstein, Teboul, Zhong, Jin, Yang, Cummings, Carvill, Shepard, McPhie,
  Torres, Ginsburg, Wang, Wu, U, Saxena, Khandelwal, Zand, Matosich, Veeraraghavan, Michelena, Li, Jagadeesh, Huang, Chawla, Huang, Chen, Garg, A, Silva, Bell, Zhang, Guo, Yu, Moshkovich, Wehrstedt, Khabsa, Avalani, Bhatt, Mankus, Hasson, Lennie, Reso, Groshev, Naumov, Lathi, Keneally, Liu, Seltzer, Valko, Restrepo, Patel, Vyatskov, Samvelyan, Clark, Macey, Wang, Hermoso, Metanat, Rastegari, Bansal, Santhanam, Parks, White, Bawa, Singhal, Egebo, Usunier, Mehta, Laptev, Dong, Cheng, Chernoguz, Hart, Salpekar, Kalinli, Kent, Parekh, Saab, Balaji, Rittner, Bontrager, Roux, Dollar, Zvyagina, Ratanchandani, Yuvraj, Liang, Alao, Rodriguez, Ayub, Murthy, Nayani, Mitra, Parthasarathy, Li, Hogan, Battey, Wang, Howes, Rinott, Mehta, Siby, Bondu, Datta, Chugh, Hunt, Dhillon, Sidorov, Pan, Mahajan, Verma, Yamamoto, Ramaswamy, Lindsay, Lindsay, Feng, Lin, Zha, Patil, Shankar, Zhang, Zhang, Wang, Agarwal, Sajuyigbe, Chintala, Max, Chen, Kehoe, Satterfield, Govindaprasad, Gupta, Deng, Cho, Virk, Subramanian, Choudhury,
  Goldman, Remez, Glaser, Best, Koehler, Robinson, Li, Zhang, Matthews, Chou, Shaked, Vontimitta, Ajayi, Montanez, Mohan, Kumar, Mangla, Ionescu, Poenaru, Mihailescu, Ivanov, Li, Wang, Jiang, Bouaziz, Constable, Tang, Wu, Wang, Wu, Gao, Kleinman, Chen, Hu, Jia, Qi, Li, Zhang, Zhang, Adi, Nam, Yu, Wang, Zhao, Hao, Qian, Li, He, Rait, DeVito, Rosnbrick, Wen, Yang, Zhao, and Ma}]{grattafiori2024llama}
Aaron Grattafiori, Abhimanyu Dubey, Abhinav Jauhri, Abhinav Pandey, Abhishek Kadian, Ahmad Al-Dahle, Aiesha Letman, Akhil Mathur, Alan Schelten, et~al. 2024.
\newblock \href {https://arxiv.org/abs/2407.21783} {The llama 3 herd of models}.
\newblock \emph{Preprint}, arXiv:2407.21783.

\bibitem[{Guo and Hu(2022)}]{10.1007/s11263-022-01667-9}
Xiaojie Guo and Qiming Hu. 2022.
\newblock \href {https://doi.org/10.1007/s11263-022-01667-9} {Low-light image enhancement via breaking down the darkness}.
\newblock \emph{Int. J. Comput. Vision}, 131(1):48–66.

\bibitem[{Horwitz et~al.(2025)Horwitz, Kurer, Kahana, Amar, and Hoshen}]{horwitz2025charting}
Eliahu Horwitz, Nitzan Kurer, Jonathan Kahana, Liel Amar, and Yedid Hoshen. 2025.
\newblock \href {https://arxiv.org/abs/2503.10633} {We should chart an atlas of all the world's models}.
\newblock \emph{Preprint}, arXiv:2503.10633.

\bibitem[{Liang et~al.(2024)Liang, Rajani, Yang, Ozoani, Wu, Chen, Smith, and Zou}]{liang2024systematic}
Weixin Liang, Nazneen Rajani, Xinyu Yang, Ezinwanne Ozoani, Eric Wu, Yiqun Chen, Daniel~Scott Smith, and James Zou. 2024.
\newblock \href {https://doi.org/10.1038/s42256-024-00857-z} {Systematic analysis of 32,111 ai model cards characterizes documentation practice in ai}.
\newblock \emph{Nature Machine Intelligence}, 6(7):744--753.

\bibitem[{Lin(2004)}]{lin2004rouge}
Chin-Yew Lin. 2004.
\newblock \href {https://aclanthology.org/W04-1013/} {{ROUGE}: A package for automatic evaluation of summaries}.
\newblock In \emph{Text Summarization Branches Out}, pages 74--81, Barcelona, Spain. Association for Computational Linguistics.

\bibitem[{Liu et~al.(2024{\natexlab{a}})Liu, Li, Jin, and Diab}]{liu2024automatic}
Jiarui Liu, Wenkai Li, Zhijing Jin, and Mona Diab. 2024{\natexlab{a}}.
\newblock \href {https://doi.org/10.18653/v1/2024.naacl-long.110} {Automatic generation of model and data cards: A step towards responsible {AI}}.
\newblock In \emph{Proceedings of the 2024 Conference of the North American Chapter of the Association for Computational Linguistics: Human Language Technologies (Volume 1: Long Papers)}, pages 1975--1997, Mexico City, Mexico. Association for Computational Linguistics.

\bibitem[{Liu et~al.(2024{\natexlab{b}})Liu, Lin, Hewitt, Paranjape, Bevilacqua, Petroni, and Liang}]{liu2024lost}
Nelson~F. Liu, Kevin Lin, John Hewitt, Ashwin Paranjape, Michele Bevilacqua, Fabio Petroni, and Percy Liang. 2024{\natexlab{b}}.
\newblock \href {https://doi.org/10.1162/tacl_a_00638} {Lost in the middle: How language models use long contexts}.
\newblock \emph{Transactions of the Association for Computational Linguistics}, 12:157--173.

\bibitem[{Longpre et~al.(2024)Longpre, Mahari, Chen, Obeng-Marnu, Sileo, Brannon, Muennighoff, Khazam, Kabbara, Perisetla et~al.}]{longpre2024large}
Shayne Longpre, Robert Mahari, Anthony Chen, Naana Obeng-Marnu, Damien Sileo, William Brannon, Niklas Muennighoff, Nathan Khazam, Jad Kabbara, et~al. 2024.
\newblock \href {https://doi.org/10.1038/s42256-024-00878-8} {A large-scale audit of dataset licensing and attribution in ai}.
\newblock \emph{Nature Machine Intelligence}, 6(8):975--987.

\bibitem[{Mitchell et~al.(2019)Mitchell, Wu, Zaldivar, Barnes, Vasserman, Hutchinson, Spitzer, Raji, and Gebru}]{mitchell2019model}
Margaret Mitchell, Simone Wu, Andrew Zaldivar, Parker Barnes, Lucy Vasserman, Ben Hutchinson, Elena Spitzer, Inioluwa~Deborah Raji, and Timnit Gebru. 2019.
\newblock \href {https://doi.org/10.1145/3287560.3287596} {Model cards for model reporting}.
\newblock In \emph{Proceedings of the Conference on Fairness, Accountability, and Transparency}, FAT* '19, page 220–229, New York, NY, USA. Association for Computing Machinery.

\bibitem[{Monroe et~al.(2008)Monroe, Colaresi, and Quinn}]{monroe2008fightin}
Burt~L Monroe, Michael~P Colaresi, and Kevin~M Quinn. 2008.
\newblock \href {https://doi.org/10.1093/pan/mpn018} {Fightin'words: Lexical feature selection and evaluation for identifying the content of political conflict}.
\newblock \emph{Political Analysis}, 16(4):372--403.

\bibitem[{NVIDIA et~al.(2025)NVIDIA, :, Blakeman, Grattafiori, Basant, Gupta, Khattar, Renduchintala, Vavre, Shukla, Bercovich, Ficek, Shaposhnikov, Kondratenko, Bukharin, Milesi, Taghibakhshi, Liu, Barton, Mahabaleshwarkar, Klein, Zuker, Geifman, Shen, Bhiwandiwalla, Tao, Guan, Mandarwal, Mehta, Aithal, Poojary, Ahamed, Thekkumpate, Dattagupta, Zhu, Sadeghi, Simkin, Lanir, Schifferer, Nushi, Kartal, Rouhani, Ginsburg, Norick, Soubasis, Kisacanin, Yu, Catanzaro, del Mundo, Hwang, Wang, Hsieh, Zhang, Yu, Mungekar, Patel, Alexiuk, Parisien, Neale, Mosk-Aoyama, Su, Corneil, Afrimi, Rohrer, Serebrenik, Gitman, Levy, Stosic, Mosallanezhad, Narayanan, Nathawani, Rekesh, Yared, Kakwani, Ahn, Riach, Stosic, Minasyan, Lin, Long, Long, Lantz, Evans, Ning, Chung, Harper, Tramel, Galinkin, Pounds, Briones, Bakhturina, Ladhak, Wang, Jia, Soares, Chen, Galko, Siino, Agam, Ajjanagadde, Bhatt, Prasad, Armstrong, Shen, Batmaz, Nalbandyan, Qian, Sharma, Ross, Ngo, Sahota, Wang, Soni, Upadhyay, Mao, Nguyen, Nguyen, Cunningham,
  Shahaf, Gitman, Loshchilov, Moshkov, Putterman, Kautz, Scowcroft, Casper, Mitra, Glick, Chen, Oliver, Zhang, Zeng, Lou, Zhang, Huang, Conway, Guman, Kamalu, Greco, Cohen, Jennings, Daw, Vialard, Yi, Parmar, Xu, Zhu, Briski, Cheung, Luna, Santhanam, Shih, Kong, Bhardwaj, Puvvada, Pawelec, Anik, McAfee, Sleiman, Derczynski, Ding, Liebenwein, Vega, Grover, Segbroeck, de~Melo, Sreedhar, Kilaru, Ashkenazi, Romeijn, Cai, Kliegl, Moosaei, Novikov, Samadi, Corpuz, Wang, Price, Boone, Evans, Martinez, Chrzanowski, Shoeybi, Patwary, Mulepati, Hereth, Assaf, Habibi, Zmora, Haber, Sessions, Bhatia, Jukar, Pope, Ludwig, Tajbakhsh, Juluru, Hrinchuk, Kuchaiev, Delalleau, Olabiyi, Argov, Xie, Chadha, Shamis, Molchanov, Morkisz, Dykas, Jin, Xu, Januszewski, Thombre, Varshney, Gundecha, Miao, Mahabadi, El-Yaniv, Zilberstein, Shafipour, Harang, Izzo, Shahbazyan, Garg, Borkar, Gala, Islam, Waleffe, Watve, Koren, Zhang, Hewett, Prenger, Timbrook, Mahdavi, Modi, Kriman, Kariyappa, Satheesh, Kaji, Pasumarthi, Narentharen,
  Narenthiran, Bak, Kashirsky, Poulos, Mor, Ramasamy, Acharya, Ghosh, Sreenivas, Thomas, Fan, Gopal, Prabhumoye, Pachori, Toshniwal, Ding, Singh, Sun, Ithape, Majumdar, Singhal, Alborghetti, Ge, Devare, Barua, Panguluri, Gupta, Priyadarshi, Akter, Bui, Ene, Kong, Do, Blankevoort, Balough, Asida, Natan, Konuk, Vashishth, Karpas, De, Noorozi, Noroozi, Srinivasan, Elango, Korthikanti, Kurin, Lavrukhin, Jiang, Ahmad, Du, Ping, Zhou, Jennings, Zhang, Prazuch, Ren, Karnati, Choi, Meyer, Wu, Zhang, Lin, Geifman, Fu, Subara, Suhara, Gao, Moshe, Dong, Liu, Chen, and Yan}]{nvidia_nemotron_nano_v3_2025}
NVIDIA, :, Aaron Blakeman, Aaron Grattafiori, Aarti Basant, Abhibha Gupta, Abhinav Khattar, Adi Renduchintala, Aditya Vavre, et~al. 2025.
\newblock \href {https://arxiv.org/abs/2512.20848} {Nemotron 3 nano: Open, efficient mixture-of-experts hybrid mamba-transformer model for agentic reasoning}.
\newblock \emph{Preprint}, arXiv:2512.20848.

\bibitem[{Olmo et~al.(2025)Olmo, :, Ettinger, Bertsch, Kuehl, Graham, Heineman, Groeneveld, Brahman, Timbers, Ivison, Morrison, Poznanski, Lo, Soldaini, Jordan, Chen, Noukhovitch, Lambert, Walsh, Dasigi, Berry, Malik, Shah, Geng, Arora, Gupta, Anderson, Xiao, Murray, Romero, Graf, Asai, Bhagia, Wettig, Liu, Rangapur, Anastasiades, Huang, Schwenk, Trivedi, Magnusson, Lochner, Liu, Miranda, Sap, Morgan, Schmitz, Guerquin, Wilson, Huff, Bras, Xin, Shao, Skjonsberg, Shen, Li, Wilde, Pyatkin, Merrill, Chang, Gu, Zeng, Sabharwal, Zettlemoyer, Koh, Farhadi, Smith, and Hajishirzi}]{olmo2025olmo}
Team Olmo, :, Allyson Ettinger, Amanda Bertsch, Bailey Kuehl, David Graham, David Heineman, Dirk Groeneveld, Faeze Brahman, et~al. 2025.
\newblock \href {https://arxiv.org/abs/2512.13961} {Olmo 3}.
\newblock \emph{Preprint}, arXiv:2512.13961.

\bibitem[{OpenAI et~al.(2025)OpenAI, :, Agarwal, Ahmad, Ai, Altman, Applebaum, Arbus, Arora, Bai, Baker, Bao, Barak, Bennett, Bertao, Brett, Brevdo, Brockman, Bubeck, Chang, Chen, Chen, Cheung, Clark, Cook, Dukhan, Dvorak, Fives, Fomenko, Garipov, Georgiev, Glaese, Gogineni, Goucher, Gross, Guzman, Hallman, Hehir, Heidecke, Helyar, Hu, Huet, Huh, Jain, Johnson, Koch, Kofman, Kundel, Kwon, Kyrylov, Le, Leclerc, Lennon, Lessans, Lezcano-Casado, Li, Li, Lin, Liss, Lily, Liu, Liu, Lu, Lu, Martinovic, McCallum, McGrath, McKinney, McLaughlin, Mei, Mostovoy, Mu, Myles, Neitz, Nichol, Pachocki, Paino, Palmie, Pantuliano, Parascandolo, Park, Pathak, Paz, Peran, Pimenov, Pokrass, Proehl, Qiu, Raila, Raso, Ren, Richardson, Robinson, Rotsted, Salman, Sanjeev, Schwarzer, Sculley, Sikchi, Simon, Singhal, Song, Stuckey, Sun, Tillet, Toizer, Tsimpourlas, Vyas, Wallace, Wang, Wang, Watkins, Weil, Wendling, Whinnery, Whitney, Wong, Yang, Yang, Yasunaga, Ying, Zaremba, Zhan, Zhang, Zhang, Zhang, and Zhao}]{agarwal2025gpt}
OpenAI, :, Sandhini Agarwal, Lama Ahmad, Jason Ai, Sam Altman, Andy Applebaum, Edwin Arbus, Rahul~K. Arora, et~al. 2025.
\newblock \href {https://arxiv.org/abs/2508.10925} {gpt-oss-120b \& gpt-oss-20b model card}.
\newblock \emph{Preprint}, arXiv:2508.10925.

\bibitem[{OpenAI et~al.(2024)OpenAI, :, Hurst, Lerer, Goucher, Perelman, Ramesh, Clark, Ostrow, Welihinda, Hayes, Radford, Mądry, Baker-Whitcomb, Beutel, Borzunov, Carney, Chow, Kirillov, Nichol, Paino, Renzin, Passos, Kirillov, Christakis, Conneau, Kamali, Jabri, Moyer, Tam, Crookes, Tootoochian, Tootoonchian, Kumar, Vallone, Karpathy, Braunstein, Cann, Codispoti, Galu, Kondrich, Tulloch, Mishchenko, Baek, Jiang, Pelisse, Woodford, Gosalia, Dhar, Pantuliano, Nayak, Oliver, Zoph, Ghorbani, Leimberger, Rossen, Sokolowsky, Wang, Zweig, Hoover, Samic, McGrew, Spero, Giertler, Cheng, Lightcap, Walkin, Quinn, Guarraci, Hsu, Kellogg, Eastman, Lugaresi, Wainwright, Bassin, Hudson, Chu, Nelson, Li, Shern, Conger, Barette, Voss, Ding, Lu, Zhang, Beaumont, Hallacy, Koch, Gibson, Kim, Choi, McLeavey, Hesse, Fischer, Winter, Czarnecki, Jarvis, Wei, Koumouzelis, Sherburn, Kappler, Levin, Levy, Carr, Farhi, Mely, Robinson, Sasaki, Jin, Valladares, Tsipras, Li, Nguyen, Findlay, Oiwoh, Wong, Asdar, Proehl, Yang, Antonow,
  Kramer, Peterson, Sigler, Wallace, Brevdo, Mays, Khorasani, Such, Raso, Zhang, von Lohmann, Sulit, Goh, Oden, Salmon, Starace, Brockman, Salman, Bao, Hu, Wong, Wang, Schmidt, Whitney, Jun, Kirchner, de~Oliveira~Pinto, Ren, Chang, Chung, Kivlichan, O'Connell, O'Connell, Osband, Silber, Sohl, Okuyucu, Lan, Kostrikov, Sutskever, Kanitscheider, Gulrajani, Coxon, Menick, Pachocki, Aung, Betker, Crooks, Lennon, Kiros, Leike, Park, Kwon, Phang, Teplitz, Wei, Wolfe, Chen, Harris, Varavva, Lee, Shieh, Lin, Yu, Weng, Tang, Yu, Jang, Candela, Beutler, Landers, Parish, Heidecke, Schulman, Lachman, McKay, Uesato, Ward, Kim, Huizinga, Sitkin, Kraaijeveld, Gross, Kaplan, Snyder, Achiam, Jiao, Lee, Zhuang, Harriman, Fricke, Hayashi, Singhal, Shi, Karthik, Wood, Rimbach, Hsu, Nguyen, Gu-Lemberg, Button, Liu, Howe, Muthukumar, Luther, Ahmad, Kai, Itow, Workman, Pathak, Chen, Jing, Guy, Fedus, Zhou, Mamitsuka, Weng, McCallum, Held, Ouyang, Feuvrier, Zhang, Kondraciuk, Kaiser, Hewitt, Metz, Doshi, Aflak, Simens, Boyd,
  Thompson, Dukhan, Chen, Gray, Hudnall, Zhang, Aljubeh, Litwin, Zeng, Johnson, Shetty, Gupta, Shah, Yatbaz, Yang, Zhong, Glaese, Chen, Janner, Lampe, Petrov, Wu, Wang, Fradin, Pokrass, Castro, de~Castro, Pavlov, Brundage, Wang, Khan, Murati, Bavarian, Lin, Yesildal, Soto, Gimelshein, Cone, Staudacher, Summers, LaFontaine, Chowdhury, Ryder, Stathas, Turley, Tezak, Felix, Kudige, Keskar, Deutsch, Bundick, Puckett, Nachum, Okelola, Boiko, Murk, Jaffe, Watkins, Godement, Campbell-Moore, Chao, McMillan, Belov, Su, Bak, Bakkum, Deng, Dolan, Hoeschele, Welinder, Tillet, Pronin, Tillet, Dhariwal, Yuan, Dias, Lim, Arora, Troll, Lin, Lopes, Puri, Miyara, Leike, Gaubert, Zamani, Wang, Donnelly, Honsby, Smith, Sahai, Ramchandani, Huet, Carmichael, Zellers, Chen, Chen, Nigmatullin, Cheu, Jain, Altman, Schoenholz, Toizer, Miserendino, Agarwal, Culver, Ethersmith, Gray, Grove, Metzger, Hermani, Jain, Zhao, Wu, Jomoto, Wu, Shuaiqi, Xia, Phene, Papay, Narayanan, Coffey, Lee, Hall, Balaji, Broda, Stramer, Xu, Gogineni,
  Christianson, Sanders, Patwardhan, Cunninghman, Degry, Dimson, Raoux, Shadwell, Zheng, Underwood, Markov, Sherbakov, Rubin, Stasi, Kaftan, Heywood, Peterson, Walters, Eloundou, Qi, Moeller, Monaco, Kuo, Fomenko, Chang, Zheng, Zhou, Manassra, Sheu, Zaremba, Patil, Qian, Kim, Cheng, Zhang, He, Zhang, Jin, Dai, and Malkov}]{hurst2024gpt}
OpenAI, :, Aaron Hurst, Adam Lerer, Adam~P. Goucher, Adam Perelman, Aditya Ramesh, Aidan Clark, AJ~Ostrow, et~al. 2024.
\newblock \href {https://arxiv.org/abs/2410.21276} {Gpt-4o system card}.
\newblock \emph{Preprint}, arXiv:2410.21276.

\bibitem[{Poznanski et~al.(2025)Poznanski, Rangapur, Borchardt, Dunkelberger, Huff, Lin, Rangapur, Wilhelm, Lo, and Soldaini}]{poznanski2025olmocr}
Jake Poznanski, Aman Rangapur, Jon Borchardt, Jason Dunkelberger, Regan Huff, Daniel Lin, Aman Rangapur, Christopher Wilhelm, Kyle Lo, and Luca Soldaini. 2025.
\newblock \href {https://arxiv.org/abs/2502.18443} {olmocr: Unlocking trillions of tokens in pdfs with vision language models}.
\newblock \emph{Preprint}, arXiv:2502.18443.

\bibitem[{Pushkarna et~al.(2022)Pushkarna, Zaldivar, and Kjartansson}]{pushkarna2022data}
Mahima Pushkarna, Andrew Zaldivar, and Oddur Kjartansson. 2022.
\newblock \href {https://doi.org/10.1145/3531146.3533231} {Data cards: Purposeful and transparent dataset documentation for responsible ai}.
\newblock In \emph{Proceedings of the 2022 ACM Conference on Fairness, Accountability, and Transparency}, FAccT '22, page 1776–1826, New York, NY, USA. Association for Computing Machinery.

\bibitem[{Rahman et~al.(2025)Rahman, Gao, and Ji}]{rahman2025hugginggraph}
Mohammad~Shahedur Rahman, Peng Gao, and Yuede Ji. 2025.
\newblock \href {https://doi.org/10.1145/3746252.3761510} {Hugginggraph: Understanding the supply chain of llm ecosystem}.
\newblock In \emph{Proceedings of the 34th ACM International Conference on Information and Knowledge Management}, CIKM '25, page 5997–6005, New York, NY, USA. Association for Computing Machinery.

\bibitem[{Team et~al.(2025)Team, Kamath, Ferret, Pathak, Vieillard, Merhej, Perrin, Matejovicova, Ramé, Rivière, Rouillard, Mesnard, Cideron, bastien Grill, Ramos, Yvinec, Casbon, Pot, Penchev, Liu, Visin, Kenealy, Beyer, Zhai, Tsitsulin, Busa-Fekete, Feng, Sachdeva, Coleman, Gao, Mustafa, Barr, Parisotto, Tian, Eyal, Cherry, Peter, Sinopalnikov, Bhupatiraju, Agarwal, Kazemi, Malkin, Kumar, Vilar, Brusilovsky, Luo, Steiner, Friesen, Sharma, Sharma, Gilady, Goedeckemeyer, Saade, Feng, Kolesnikov, Bendebury, Abdagic, Vadi, György, Pinto, Das, Bapna, Miech, Yang, Paterson, Shenoy, Chakrabarti, Piot, Wu, Shahriari, Petrini, Chen, Lan, Choquette-Choo, Carey, Brick, Deutsch, Eisenbud, Cattle, Cheng, Paparas, Sreepathihalli, Reid, Tran, Zelle, Noland, Huizenga, Kharitonov, Liu, Amirkhanyan, Cameron, Hashemi, Klimczak-Plucińska, Singh, Mehta, Lehri, Hazimeh, Ballantyne, Szpektor, Nardini, Pouget-Abadie, Chan, Stanton, Wieting, Lai, Orbay, Fernandez, Newlan, yeong Ji, Singh, Black, Yu, Hui, Vodrahalli, Greff, Qiu,
  Valentine, Coelho, Ritter, Hoffman, Watson, Chaturvedi, Moynihan, Ma, Babar, Noy, Byrd, Roy, Momchev, Chauhan, Sachdeva, Bunyan, Botarda, Caron, Rubenstein, Culliton, Schmid, Sessa, Xu, Stanczyk, Tafti, Shivanna, Wu, Pan, Rokni, Willoughby, Vallu, Mullins, Jerome, Smoot, Girgin, Iqbal, Reddy, Sheth, Põder, Bhatnagar, Panyam, Eiger, Zhang, Liu, Yacovone, Liechty, Kalra, Evci, Misra, Roseberry, Feinberg, Kolesnikov, Han, Kwon, Chen, Chow, Zhu, Wei, Egyed, Cotruta, Giang, Kirk, Rao, Black, Babar, Lo, Moreira, Martins, Sanseviero, Gonzalez, Gleicher, Warkentin, Mirrokni, Senter, Collins, Barral, Ghahramani, Hadsell, Matias, Sculley, Petrov, Fiedel, Shazeer, Vinyals, Dean, Hassabis, Kavukcuoglu, Farabet, Buchatskaya, Alayrac, Anil, Dmitry, Lepikhin, Borgeaud, Bachem, Joulin, Andreev, Hardin, Dadashi, and Hussenot}]{gemma_2025}
Gemma Team, Aishwarya Kamath, Johan Ferret, Shreya Pathak, Nino Vieillard, Ramona Merhej, Sarah Perrin, Tatiana Matejovicova, Alexandre Ramé, et~al. 2025.
\newblock \href {https://arxiv.org/abs/2503.19786} {Gemma 3 technical report}.
\newblock \emph{Preprint}, arXiv:2503.19786.

\bibitem[{Wang et~al.(2023)Wang, Wei, Schuurmans, Le, Chi, Narang, Chowdhery, and Zhou}]{wangself}
Xuezhi Wang, Jason Wei, Dale Schuurmans, Quoc~V Le, Ed~H. Chi, Sharan Narang, Aakanksha Chowdhery, and Denny Zhou. 2023.
\newblock \href {https://openreview.net/forum?id=1PL1NIMMrw} {Self-consistency improves chain of thought reasoning in language models}.
\newblock In \emph{The Eleventh International Conference on Learning Representations}.

\bibitem[{Wolf et~al.(2020)Wolf, Debut, Sanh, Chaumond, Delangue, Moi, Cistac, Rault, Louf, Funtowicz, Davison, Shleifer, von Platen, Ma, Jernite, Plu, Xu, Scao, Gugger, Drame, Lhoest, and Rush}]{wolf2019huggingface}
Thomas Wolf, Lysandre Debut, Victor Sanh, Julien Chaumond, Clement Delangue, Anthony Moi, Pierric Cistac, Tim Rault, Rémi Louf, et~al. 2020.
\newblock \href {https://arxiv.org/abs/1910.03771} {Huggingface's transformers: State-of-the-art natural language processing}.
\newblock \emph{Preprint}, arXiv:1910.03771.

\bibitem[{Yang et~al.(2025)Yang, Li, Yang, Zhang, Hui, Zheng, Yu, Gao, Huang, Lv, Zheng, Liu, Zhou, Huang, Hu, Ge, Wei, Lin, Tang, Yang, Tu, Zhang, Yang, Yang, Zhou, Zhou, Lin, Dang, Bao, Yang, Yu, Deng, Li, Xue, Li, Zhang, Wang, Zhu, Men, Gao, Liu, Luo, Li, Tang, Yin, Ren, Wang, Zhang, Ren, Fan, Su, Zhang, Zhang, Wan, Liu, Wang, Cui, Zhang, Zhou, and Qiu}]{yang2025qwen3}
An~Yang, Anfeng Li, Baosong Yang, Beichen Zhang, Binyuan Hui, Bo~Zheng, Bowen Yu, Chang Gao, Chengen Huang, et~al. 2025.
\newblock \href {https://arxiv.org/abs/2505.09388} {Qwen3 technical report}.
\newblock \emph{Preprint}, arXiv:2505.09388.

\bibitem[{Yang et~al.(2024)Yang, Liang, and Zou}]{yangnavigating}
Xinyu Yang, Weixin Liang, and James Zou. 2024.
\newblock \href {https://openreview.net/forum?id=xC8xh2RSs2} {Navigating dataset documentations in {AI}: A large-scale analysis of dataset cards on huggingface}.
\newblock In \emph{The Twelfth International Conference on Learning Representations}.

\bibitem[{Zhang et~al.(2020)Zhang, Kishore, Wu, Weinberger, and Artzi}]{zhangbertscore}
Tianyi Zhang, Varsha Kishore, Felix Wu, Kilian~Q. Weinberger, and Yoav Artzi. 2020.
\newblock \href {https://openreview.net/forum?id=SkeHuCVFDr} {Bertscore: Evaluating text generation with bert}.
\newblock In \emph{International Conference on Learning Representations}.

\bibitem[{Zhang et~al.(2025)Zhang, Li, Long, Zhang, Lin, Yang, Xie, Yang, Liu, Lin, Huang, and Zhou}]{zhang2025qwen3}
Yanzhao Zhang, Mingxin Li, Dingkun Long, Xin Zhang, Huan Lin, Baosong Yang, Pengjun Xie, An~Yang, Dayiheng Liu, et~al. 2025.
\newblock \href {https://arxiv.org/abs/2506.05176} {Qwen3 embedding: Advancing text embedding and reranking through foundation models}.
\newblock \emph{Preprint}, arXiv:2506.05176.

\bibitem[{Zheng et~al.(2023)Zheng, Chiang, Sheng, Zhuang, Wu, Zhuang, Lin, Li, Li, Xing, Zhang, Gonzalez, and Stoica}]{zheng2023judging}
Lianmin Zheng, Wei-Lin Chiang, Ying Sheng, Siyuan Zhuang, Zhanghao Wu, Yonghao Zhuang, Zi~Lin, Zhuohan Li, Dacheng Li, et~al. 2023.
\newblock \href {https://proceedings.neurips.cc/paper_files/paper/2023/file/91f18a1287b398d378ef22505bf41832-Paper-Datasets_and_Benchmarks.pdf} {Judging llm-as-a-judge with mt-bench and chatbot arena}.
\newblock In \emph{Advances in Neural Information Processing Systems}, volume~36, pages 46595--46623. Curran Associates, Inc.

\bibitem[{Zhou et~al.(2022)Zhou, Wei, Wang, Shen, Xie, Yuille, and Kong}]{zhouimage}
Jinghao Zhou, Chen Wei, Huiyu Wang, Wei Shen, Cihang Xie, Alan Yuille, and Tao Kong. 2022.
\newblock \href {https://openreview.net/forum?id=ydopy-e6Dg} {Image {BERT} pre-training with online tokenizer}.
\newblock In \emph{International Conference on Learning Representations}.

\end{thebibliography}
\appendix

\section{Definition of Model and Data Card}
\label{sec:definition}

\begin{table*}[t]
\small
\centering
\begin{tabular}{cp{4.2cm}p{11cm}}
\toprule
\textbf{Card} & \textbf{Field} & \textbf{Description} \\
\midrule
\multirow{8}{*}[-25pt]{\rotatebox[origin=c]{90}{\textbf{Model Card}}} 
 & Model Details & Model architecture, parameter count, developer organization, license, release date, and other model metadata \\

 & Intended Use & Primary applications, target audience, out-of-scope uses, domain restrictions, and other intended uses \\

 & Generative Capabilities & Context length, inference latency, multilingual support, supported modalities, and other capabilities \\

 & Safety Considerations & Safety alignment methods, red teaming results, jailbreak resistance, harm reduction strategies, and other safety analysis \\

 & Training Data & Training corpus, data mixture, filtering pipeline, data provenance, and other training data information \\

 & Performance Metrics & Benchmark results, accuracy metrics, reasoning scores, safety scores, robustness metrics, and other quantitative metrics \\

 & Ethical Considerations & Environmental impact, intellectual property, dual-use risks, societal implications, and other ethical issues \\

 & Caveats \& Recommendations & Hardware requirements, deployment guidelines, operational constraints, model limitations, and other recommendations \\
\hline
\multirow{12}{*}[-40pt]{\rotatebox[origin=c]{90}{\textbf{Data Card}}} 
 & Dataset Details & Dataset name, version identifier, creators and curators, funding, license, text language, and other dataset metadata \\

 & Dataset Structure & Instance count, field schema, data splits, dataset size, and other structural details \\

 & Data Collection & Collection methodology, data sources, collection timeframe, consent process, and other collection details \\

 & Data Processing & Preprocessing steps, cleaning procedures, labeling process, filtering criteria, deduplication, and other processing steps \\

 & Intended Uses & Primary tasks, intended use cases, prohibited uses, commercial restrictions, and other usage information \\

 & Bias \& Fairness & Demographic distribution, geographic coverage, social bias analysis, fairness assessment, and other bias or fairness information \\

 & Privacy \& Security & Personally identifiable information (PII), anonymization methods, data security protocols, confidentiality measures, and other privacy measures \\

 & Content Analysis & Content types, toxicity analysis, misinformation risks, offensive language, and other content risks \\

 & Legal \& Ethical & Copyright status, terms of use, ethical review, compliance requirements, and other legal details \\

 & Maintenance \& Updates & Maintenance plan, update frequency, versioning policy, deprecation plan, and other maintenance information \\

 & Distribution \& Access & Access mechanism, download instructions, repository link, citation requirements, and other access details \\

 & Limitations \& Recommendations & Known limitations, usage guidelines, quality caveats, and other recommendations \\
\bottomrule
\end{tabular}
\caption{Definitions of Model and Data Card for Generative AI}
\label{tab:definition_academic}
\end{table*}

The definitions of Model card and Data card for GenAI are shown in Table \ref{tab:definition_academic}.

\section{Dataset Characteristics}
\label{appendix:statistics}

\begin{figure*}[t]
    \centering
    \includegraphics[width=\linewidth]{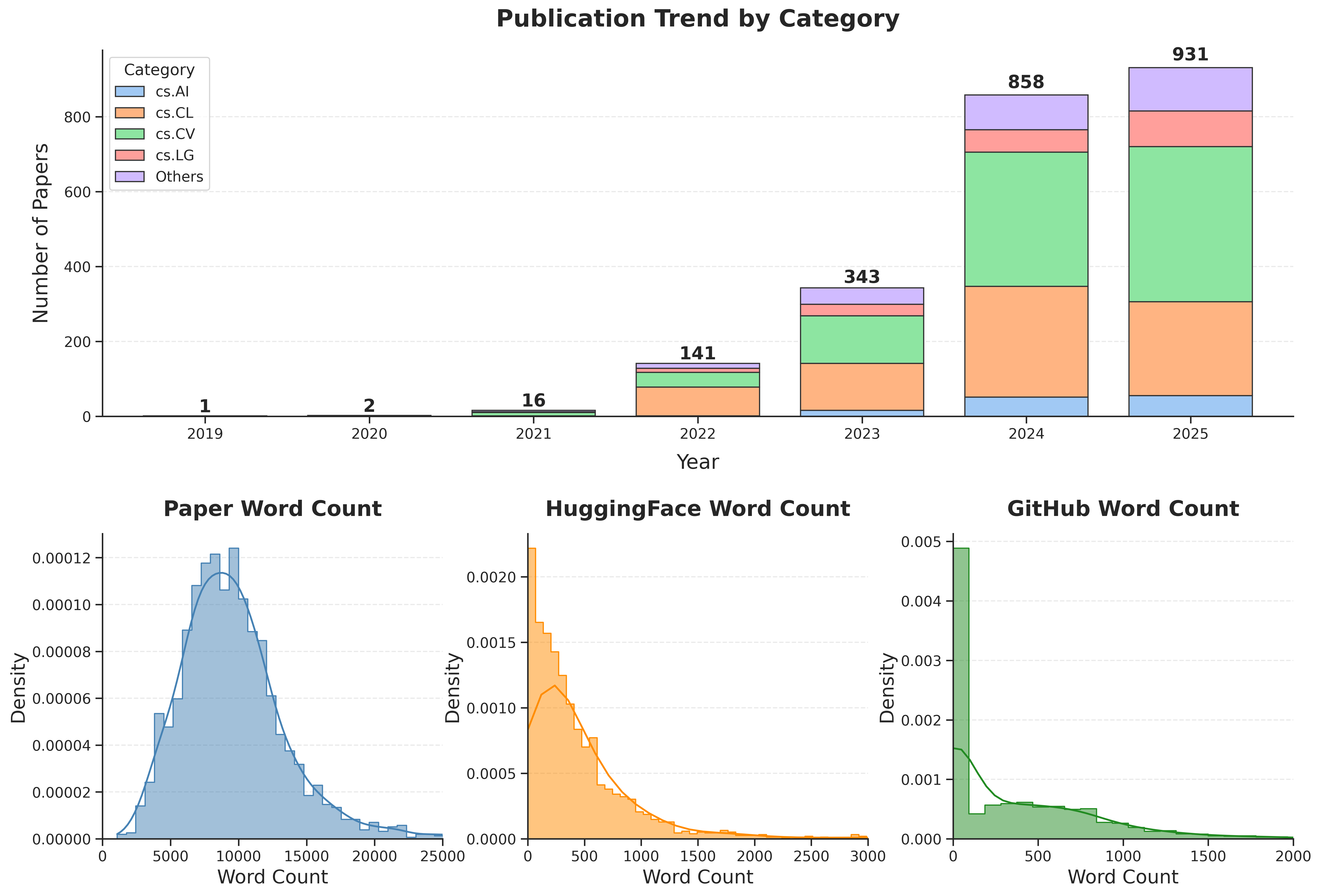}
    \caption{\textbf{Dataset Overview.} \textit{Top:} Publication trends of the 2,541 collected triplets (2019--2025), categorized by primary arXiv domain. \textit{Bottom:} Word count distributions across three data sources (Academic Papers, Hugging Face, GitHub), illustrating complementary information granularities.}
    \label{fig:data_stats}
\end{figure*}

Figure~\ref{fig:data_stats} (Top) illustrates the temporal distribution of our corpus, which shows an increase from 141 triplets in 2022 to 931 in 2025. The domain composition is dominated by Computer Vision (\texttt{cs.CV}) and Computational Linguistics (\texttt{cs.CL}), reflecting our filtering criteria targeting papers with verified GitHub and Hugging Face artifacts in generative modeling. The word count analysis (Bottom) reveals distinct content characteristics across sources: academic papers provide a comprehensive context (median $\approx$10k words), while GitHub and Hugging Face README files offer concise implementation details (median $<$1k words). This complementary information structure empirically justifies our multi-source triangulation approach.

\section{MetaGAI Benchmark Generation Algorithm}

The Algorithm~\ref{alg:metagai_pipeline} formalizes the three-stage MetaGAI generation pipeline. For each schema key $k_i \in \mathcal{K}$, the Retriever Agent maps segmented source chunks $\mathcal{X}$ to relevant evidence spans, which are passed to the Generator Agent to synthesize an independent structured draft. The Editor Agent then consolidates each draft by cross-validating against the full multi-source context $\mathcal{P} \cup \mathcal{S}$, verifying attribution, and merging complementary content to produce the final card $\hat{\mathcal{C}}$.

\begin{algorithm}[t]
\caption{MetaGAI Benchmark Generation Pipeline}
\label{alg:metagai_pipeline}
\begin{algorithmic}[1]
\Require 
    $\mathcal{P} \cup \mathcal{S}$: Full Context (Paper, GitHub, Hugging Face)
    \quad $\mathcal{X}$: Segmented chunks of $\mathcal{P} \cup \mathcal{S}$
    \quad $\mathcal{K}$: Schema Keys $\{k_i\}_{i=1}^M$
\Ensure 
    $\hat{\mathcal{C}}$: Final Card

\Statex \textbf{Stage 1: Retriever Agent (Evidence Mapping)}

\State $\mathcal{E} \gets \emptyset$
\For{each $k_i \in \mathcal{K}$}
    \State $v_{evidence} \gets \text{Retriever}(\mathcal{X}, k_i)$ \Comment{Extract factual content}
    \State $\mathcal{E}[k_i] \gets v_{evidence}$
\EndFor

\Statex \textbf{Stage 2: Generator Agent (Draft Synthesis)}

\State $\tilde{\mathcal{C}} \gets \emptyset$
\For{each $k_i \in \mathcal{K}$}
    \State $\tilde{v}_i \gets \text{Generator}(\mathcal{E}[k_i], \text{Prompt}_{\text{Gen}})$ \Comment{Synthesize structured draft}
    \State $\tilde{\mathcal{C}}[k_i] \gets \tilde{v}_i$
\EndFor

\Statex \textbf{Stage 3: Editor Agent (Consolidation)}

\For{each $k_i \in \mathcal{K}$}
    \State \Comment{Editor verifies draft against full context}
    \State $\hat{v}_i \gets \text{Editor}(\mathcal{P} \cup \mathcal{S}, \tilde{\mathcal{C}}[k_i], \text{Prompt}_{\text{Edit}})$ 
    \State $\hat{\mathcal{C}}[k_i] \gets \hat{v}_i$
\EndFor

\State \Return $\hat{\mathcal{C}}$
\end{algorithmic}
\end{algorithm}

\section{MetaGAI Quality Validation Protocols}
\label{appendix:eval_metrics}

To rigorously assess generated card quality, we employ a comprehensive evaluation framework. This framework adopts three established metrics (\textit{Faithfulness}, \textit{Relevance}, and \textit{Accuracy}) from previous work~\cite{liu2024automatic}, and introduces two additional metrics (\textit{Consistency} and \textit{Usefulness}) to address the structural and practical requirements of technical documentation.

\subsection{Metric Definitions}

\paragraph{Faithfulness (F).} 
Measures the degree to which generated content is grounded in provided source materials (paper, GitHub, Hugging Face), ensuring absence of unsupported claims or hallucinations.

\paragraph{Relevance (R).} 
Evaluates the information density and pertinence to the target field. High-scoring content strictly adheres to field definitions, avoiding redundancy, verbosity, or out-of-scope information.

\paragraph{Accuracy (A).} 
Assesses factual correctness of technical details (numerical values, licenses, entity names). Unlike Faithfulness, which verifies source alignment, Accuracy validates objective correctness against domain knowledge.

\paragraph{Consistency (C).} 
Evaluates internal logical coherence across sentences and fields within a card, detecting contradictions that compromise documentation integrity.

\paragraph{Usefulness (U).} 
Measures practical value for downstream users. Useful content provides specific, actionable insights that support deployment decisions, rather than generic descriptions.

\subsection{Scoring Rubric}

Each metric is evaluated on a 1--5 Likert scale. Table~\ref{tab:scoring_rubric} presents the scoring criteria applied by both human evaluators and LLM judges.

\begin{table*}[t]
\small
\centering
\begin{tabular}{p{2cm}p{4.5cm}p{4.5cm}p{4.5cm}}
\toprule
\textbf{Metric} & \textbf{Score 1 (Poor)} & \textbf{Score 3 (Acceptable)} & \textbf{Score 5 (Excellent)} \\
\midrule
\textbf{Faithfulness} & 
Contains major hallucinations; claims unsupported by sources. & 
Mostly supported with minor extrapolations. & 
Fully grounded in source text; no unsupported claims. \\
\addlinespace
\textbf{Relevance} & 
Largely off-topic, redundant, or contains significant noise. & 
Generally on-topic with some redundant details. & 
Concise and strictly focused on field definition. \\
\addlinespace
\textbf{Accuracy} & 
Contains critical factual errors. & 
Core facts correct; minor imprecisions present. & 
Factually precise; all technical details correct. \\
\addlinespace
\textbf{Consistency} & 
Contains direct logical contradictions. & 
Generally consistent; minor ambiguities present. & 
Logically coherent; no internal contradictions. \\
\addlinespace
\textbf{Usefulness} & 
Vague or generic; provides no practical value. & 
Provides basic information but lacks depth. & 
Rich, actionable insights supporting deployment. \\
\bottomrule
\end{tabular}
\caption{\textbf{Evaluation Scoring Rubric.} Criteria for assessing generated metadata cards on a 1--5 Likert scale across five quality dimensions.}
\label{tab:scoring_rubric}
\end{table*}

\subsection{Retrieval Strategy Validation: Experimental Details}
\label{appendix:retriever_protocol}

Ten randomly sampled entries were segmented and evaluated by two domain experts. From 6,116 chunk-field decisions, we retained 276 chunks achieving unanimous consensus as ground truth. We compared BGE-Reranker-v2-m3~\cite{chen2024bge}, Qwen3-Reranker-8B~\cite{zhang2025qwen3}, Llama-3.1-8B-Instruct~\cite{grattafiori2024llama}, Qwen2.5-7B-Instruct, and Qwen3-30B-A3B-Instruct~\cite{yang2025qwen3}. We report Precision at rank 1 (P@1) to measure top-result accuracy, alongside Recall and F1 at rank 5 (R@5, F1@5) to assess broader retrieval coverage. The results are shown in Table~\ref{tab:retriever_val}.

\subsection{Human Annotation Protocol for Benchmark Quality Assessment}
\label{appendix:humaneval_protocol}

\paragraph{Annotators and Sample.}
Two Ph.D.\ students specializing in NLP (Ann-1, Ann-2) independently annotated a stratified random sample of 100 cards (50 Model Cards, 50 Data Cards). The cards were presented in random order under blind conditions.

\paragraph{Evaluation Instrument.}
Each card was scored on a 1--5 Likert scale in five quality dimensions defined in Appendix~\ref{appendix:eval_metrics}: Faithfulness, Relevance, Accuracy, Consistency, and Usefulness. Annotators were provided with detailed rubrics and calibration examples prior to the main annotation task.

\paragraph{LLM Judge Configuration.}
Three LLM judges --Llama-3.3-70B-Instruct~\cite{grattafiori2024llama}, GPT-OSS-120B~\cite{agarwal2025gpt}, and Qwen3-235B-A22B-2507~\cite{yang2025qwen3} -- evaluated the same outputs under randomized presentation order. Each judge was queried three times per instance with independently shuffled orderings; scores were averaged across runs to mitigate position bias.

\begin{table}[ht]
\centering
\small
\setlength{\tabcolsep}{4pt}
\begin{tabular}{lcc}
\toprule
\textbf{Target Judge} & \textbf{Ann-1} & \textbf{Ann-2} \\
\midrule
Ann-2              & $0.439^{*}$ & --- \\
\midrule
Llama-3.3-70B      & $0.675^{*}$ & $0.493^{*}$ \\
GPT-OSS-120B       & $0.087$\ (n.s.) & $0.191^{*}$ \\
Qwen3-235B         & $0.703^{*}$ & $0.367^{*}$ \\
\bottomrule
\end{tabular}
\caption{Instance-level Pearson correlations between domain annotators and LLM judges. $^{*}p < 0.001$; n.s.\ $p \geq 0.05$.}
\label{tab:pearson_agreement}
\end{table}

\paragraph{Inter-Annotator and Human-LLM Agreement.}
Table~\ref{tab:pearson_agreement} reports Pearson correlation coefficients at the instance level (per card $\times$ system pair). The human inter-annotator agreement is moderate ($r = 0.44$, $p < 0.001$), consistent with the documented subjectivity of documentation quality judgments. Among LLM judges, Qwen3-235B and Llama-3.3-70B achieve a strong and significant alignment with Ann-1 ($r = 0.70$ and $r = 0.67$; both $p < 0.001$) and moderate significant alignment with Ann-2 ($r = 0.37$ and $r = 0.49$; both $p < 0.001$). GPT-OSS-120B does not achieve a significant correlation with Ann-1 ($r = 0.09$, $p = 0.054$, n.s.) and shows only weak agreement with Ann-2 ($r = 0.19$, $p < 0.001$), motivating the exclusion of any single-judge design and supporting our ensemble evaluation strategy.

\section{Inter-Judge Agreement Analysis}
\label{appendix:judge_agreement}

To validate the robustness of our LLM-as-a-Judge framework, we computed Pearson correlation coefficients between the three judge models in the stratified evaluation set ($N=500$). Table~\ref{tab:judge_correlations} reveals substantial variation between judges. While Qwen3-235B-A22B-2507 and Llama-3.3-70B-Instruct demonstrate moderate alignment ($\rho=0.540$), GPT-OSS-120B exhibits markedly different scoring patterns (correlations $\leq 0.226$). All observed correlations achieve statistical significance ($p < 0.001$), indicating that the divergence of the score reflects systematic architectural differences in quality assessment rather than random measurement errors. This architectural diversity justifies our ensemble averaging strategy: aggregating judgments across models with distinct evaluation perspectives mitigates single-model biases and provides more robust quality estimates than any individual judge.

\begin{table}[t]
    \centering
    \small 
    \setlength{\tabcolsep}{4pt}
    \begin{tabular}{l c c c}
        \toprule
        \textbf{Judge Model} & \textbf{Llama} & \textbf{GPT} & \textbf{Qwen} \\
        \midrule
        Llama-3.3-70B-Instruct & 1.000 & 0.101$^{*}$ & 0.540$^{*}$ \\
        GPT-OSS-120B           & 0.101$^{*}$ & 1.000 & 0.226$^{*}$ \\
        Qwen3-235B-A22B-2507   & 0.540$^{*}$ & 0.226$^{*}$ & 1.000 \\
        \bottomrule
    \end{tabular}
    \caption{\textbf{Inter-Judge Correlation Matrix.} Pearson correlations computed on 500 samples. All correlations are statistically significant ($^{*}p < 0.001$). \textbf{Llama}: Llama-3.3-70B-Instruct, \textbf{GPT}: GPT-OSS-120B, \textbf{Qwen}: Qwen3-235B-A22B-2507.}
    \label{tab:judge_correlations}
\end{table}

\section{Prompt Templates}
\label{appendix:prompts}

\subsection{Semantic Verification Prompt}
\label{appendix:prompt_correspondence}

{\small
\begin{tcolorbox}[
    colback=blue!5, 
    colframe=blue!50!black, 
    boxrule=1pt, 
    rounded corners, 
    title=\textbf{Semantic Verification Prompt}, 
    fonttitle=\bfseries,
    label={prompt:verification}
]

\textbf{System Instruction}\\
You are an expert research evaluator specializing in [dataset/model] identification.

\vspace{4pt}
\textbf{Task Description} \\
Determine whether a research paper introduction and two README files (GitHub and Hugging Face) describe the \textbf{same} [dataset/model].

\vspace{4pt}
\textbf{Evaluation Criteria}

\begin{enumerate}[leftmargin=1.5em, itemsep=0pt, parsep=0pt, topsep=2pt]
    \item \textbf{Existence}: Does the paper explicitly introduce a specific named entity?
    \item \textbf{Consistency}: Do BOTH the GitHub README and Hugging Face README reference the SAME entity as the paper?
    \item \textbf{Alignment}: Compare names, domain, size metrics, methodology, and unique features.
\end{enumerate}

\vspace{4pt}
\textbf{Input Data} \\
\textbf{Paper Content}: \textit{<Paper text>}\\
\textbf{GitHub README}: \textit{<GitHub README text>}\\
\textbf{Hugging Face README}: \textit{<HF README text>}

\vspace{4pt}
\textbf{Output Format} \\
RELATED: [Yes/No]\\
CONFIDENCE: [High/Medium/Low]\\
EXPLANATION: [2-3 sentences citing specific evidence]

\end{tcolorbox}
}

\subsection{Retriever Agent Prompt}
\label{appendix:prompt_retriever}

{\small
\begin{tcolorbox}[
    colback=green!5, 
    colframe=green!50!black, 
    boxrule=1pt, 
    rounded corners, 
    title=\textbf{Retriever Agent Prompt}, 
    fonttitle=\bfseries,
    label={prompt:retriever}
]

\textbf{System Instruction}\\
You are a metadata classification assistant. Output valid JSON only. STRICTLY select sub-fields from the provided ontology.

\vspace{4pt}
\textbf{Task Description} \\
Classify the provided text chunk into predefined metadata fields for [Dataset/Model] documentation.

\vspace{4pt}
\textbf{Evaluation Criteria}
\begin{enumerate}[leftmargin=1.5em, itemsep=0pt, parsep=0pt, topsep=2pt]
    \item \textbf{Relevance}: Assign a score (0--4) indicating how well the chunk describes the field.
    \item \textbf{Sub-field Selection}: Select 1--5 sub-fields \textbf{strictly} from the provided ``Selectable Sub-fields'' list. Do not extract raw text.
\end{enumerate}

\vspace{4pt}
\textbf{Input Data} \\
\textbf{Source}: \textit{<[PAPER] / [GITHUB] / [HUGGINGFACE]>}\\
\textbf{Content}: \textit{<Text chunk content>}

\vspace{4pt}
\textbf{Output Format} \\
\texttt{\{
  "classifications": [
   \{\\ "field": "field\_name", \\
    "relevance": 3, \\
    "matched\_sub\_fields": ["sub\_field1"] \}
  ]
\}}

\end{tcolorbox}
}

\subsection{Generator Agent Prompt}
\label{appendix:prompt_generator}

{\small
\begin{tcolorbox}[
    colback=green!5, 
    colframe=green!50!black, 
    boxrule=1pt, 
    rounded corners, 
    title=\textbf{Generator Agent Prompt}, 
    fonttitle=\bfseries,
    label={prompt:generator}
]

\textbf{System Instruction}\\
You are an expert AI Researcher. Generate specific metadata sections for a [Model/Dataset] Card. Output valid JSON only.

\vspace{4pt}
\textbf{Critical Constraint: Subject Focus}
\begin{itemize}[leftmargin=1.5em, itemsep=0pt, parsep=0pt, topsep=2pt]
    \item \textbf{Focus}: Solely on the entity \textbf{introduced} in this paper.
    \item \textbf{Ignore}: Baselines, pre-training models, or comparisons.
\end{itemize}

\vspace{4pt}
\textbf{Input Data} \\
\textbf{Target}: \textit{<Field Name>} (e.g., \textit{Training Data}) $\mid$ \textbf{Context}: \textit{<Retrieved chunks>}

\vspace{4pt}
\textbf{Output Requirements (Per Sub-field)}
\begin{enumerate}[leftmargin=1.5em, itemsep=0pt, parsep=0pt, topsep=2pt]
    \item \textbf{Content}: Summarized factual answer.
    \item \textbf{Evidence Quote}: Direct verbatim quote supporting the answer.
    \item \textbf{Confidence}: [low, medium, high, certain].
    \item \textbf{Source}: Provenance (e.g., ``Paper+GitHub'').
\end{enumerate}

\vspace{4pt}
\textbf{Output Format} \\
\texttt{\{ "sub\_field": \{ \\"content": "...", \\"evidence\_quote": "...", \\"confidence": "high", \\"source": "Paper" \} \}}

\end{tcolorbox}
}

\subsection{Editor Agent Prompt}
\label{appendix:prompt_editor}

{\small
\begin{tcolorbox}[
    colback=green!5, 
    colframe=green!50!black, 
    boxrule=1pt, 
    rounded corners, 
    title=\textbf{Editor Agent Prompt}, 
    fonttitle=\bfseries,
    label={prompt:editor}
]

\textbf{System Instruction}\\
You are the Chief Editor. Consolidate 3 candidate drafts into ONE \textbf{concise} entry describing the \textbf{proposed} entity.

\vspace{4pt}
\textbf{Task Description} \\
Review three independent drafts (Candidates A, B, C) against the source ground truth. Identify the correct \textbf{proposed} entity, filter out hallucinations, and merge valid details.

\vspace{4pt}
\textbf{Evaluation Logic}
\begin{enumerate}[leftmargin=1.5em, itemsep=0pt, parsep=0pt, topsep=2pt]
    \item \textbf{Identify Proposed Entity}: Describe ONLY the entity introduced in THIS paper. Discard candidates describing baselines.
    \item \textbf{Verify Evidence}: Check attribution.
    \item \textbf{Conciseness}: Remove fluff; use direct facts (1--3 sentences).
\end{enumerate}

\vspace{4pt}
\textbf{Input Data} \\
\textbf{Target}: \textit{<Field Name>} $\mid$ \textbf{Ground Truth}: \textit{<Raw chunks>}\\
\textbf{Candidates}: \textit{<Draft A>}, \textit{<Draft B>}, \textit{<Draft C>}

\vspace{4pt}
\textbf{Output Format} \\
\texttt{\{
  \\"selected\_candidates": ["Candidate B"],
 \\ "final\_content": "...",
 \\ "final\_evidence": "...",
 \\ "reasoning": "..."
\}}

\end{tcolorbox}
}

\subsection{Evaluation Judge Prompt}
\label{appendix:prompt_judge}

{\small
\begin{tcolorbox}[
    colback=yellow!5, 
    colframe=yellow!50!black, 
    boxrule=1pt, 
    rounded corners, 
    title=\textbf{Evaluation Judge Prompt}, 
    fonttitle=\bfseries,
    label={prompt:judge}
]

\textbf{System Instruction} \\
You are an expert evaluator for Generative AI documentation. Your task is to compare the generated metadata card against the provided triangulated sources (Paper, GitHub, Hugging Face).

\vspace{4pt}
\textbf{Task Description} \\
Evaluate the candidate text based on the five dimensions below. Assign a score from 1 (Poor) to 5 (Excellent) for each dimension.

\vspace{4pt}
\textbf{Evaluation Metrics}
\begin{itemize}[leftmargin=1.5em, itemsep=0pt, parsep=0pt, topsep=2pt]
    \item \textbf{Faithfulness (F)}
    \item \textbf{Relevance (R)}
    \item \textbf{Accuracy (A)}
    \item \textbf{Consistency (C)}
    \item \textbf{Usefulness (U)}
\end{itemize}

\vspace{4pt}
\textbf{Analysis Considerations} \\
Classify content into: Accurate facts (verifiable), Vague facts, Logical reasoning, Illogical reasoning, or Acceptable inferences.

\vspace{4pt}
\textbf{Input Data} \\
\texttt{Sources: <Context>; Target Field: <Field Name>; Candidate: <Text>}

\vspace{4pt}
\textbf{Output Format} \\
\texttt{\{"scores": \{"F": 5, "R": 5, "A": 5, "C": 5, "U": 5\}, "reasoning": "..."\}}

\end{tcolorbox}
}

\section{Comprehensive Analysis of Benchmark Characteristics and Baseline Behaviors}
\label{appendix:comprehensive_analysis}

To understand the MetaGAI benchmark's inherent challenges and model behaviors, we conduct a granular analysis covering information provenance, information sparsity impact, comparative lexical quality, and cost-efficiency trade-offs.

\subsection{Information Provenance and Editor Dynamics}
\label{sec:provenance_dynamics}

A central design principle of MetaGAI is that high-fidelity card generation requires multi-source triangulation. We validate this through provenance analysis of final ground-truth fields.

\paragraph{Source Distribution.}
Table~\ref{tab:source_provenance} demonstrates that while academic papers provide the majority of information (45--66\%), a substantial proportion of metadata fields (22--31\%) requires multi-source triangulation across paper, GitHub, and Hugging Face artifacts. Data Cards exhibit higher reliance on multi-source integration (27--31\%) compared to Model Cards (22\%), confirming that dataset documentation is typically fragmented across theoretical descriptions in papers and implementation details in repositories, necessitating cross-source synthesis.

\begin{table}[t]
    \centering
    \small
    \begin{tabular}{l c c c c}
        \toprule
        \multirow{2}{*}{\textbf{Source Category}} & \multicolumn{2}{c}{\textbf{Bench\_Mistral}} & \multicolumn{2}{c}{\textbf{Bench\_GPT-OSS}} \\
        \cmidrule(lr){2-3} \cmidrule(lr){4-5}
         & \textbf{Model} & \textbf{Data} & \textbf{Model} & \textbf{Data} \\
        \midrule
        Paper Only & 65.9\% & 61.9\% & 57.1\% & 45.1\% \\
        Multi-Source & 22.4\% & 26.7\% & 22.2\% & 30.7\% \\
        GitHub Only & 3.5\% & 3.3\% & 3.4\% & 3.9\% \\
        Not Provided & 8.2\% & 8.1\% & 17.4\% & 20.3\% \\
        \bottomrule
    \end{tabular}
    \caption{\textbf{Information Provenance Distribution.} Percentage of metadata fields derived from single vs. multiple sources across the two benchmark subsets.}
    \label{tab:source_provenance}
\end{table}

\paragraph{Editor Strategy.}
Table~\ref{tab:editor_behavior} reveals different consolidation patterns. The Best Match strategy dominates (80--95\%), while the Merge strategy contributes 5--20\% depending on the editor architecture. This demonstrates that achieving comprehensive ground truth for certain fields necessitates synthesizing complementary information across multiple generator drafts rather than selecting a single best candidate.

\begin{table}[t]
    \centering
    \small
    \begin{tabular}{l c c}
        \toprule
     & \textbf{Bench\_Mistral} & \textbf{Bench\_GPT-OSS} \\
        \midrule
        \multicolumn{3}{l}{\textit{Consolidation Strategy}} \\
        \quad Best Match & 80.0\% & 94.9\% \\
        \quad Merge & 20.0\% & 5.1\% \\
        \midrule
        \multicolumn{3}{l}{\textit{Winning Candidate Source}} \\
        \quad Qwen2.5-7B & 64.6\% & 63.8\% \\
        \quad Olmo3-7B & 27.8\% & 30.3\% \\
        \quad Llama-3.1-8B & 7.6\% & 5.9\% \\
        \bottomrule
    \end{tabular}
    \caption{\textbf{Editor Agent Dynamics.} \textit{Top:} The logic used by the Editor to finalize content. \textit{Bottom:} The distribution of Generator drafts selected as the final ground truth.}
    \label{tab:editor_behavior}
\end{table}

\subsection{Ablation Study: Source Completeness vs.\ Model Inferential Limits}
\label{sec:ablation_provenance}

The faithfulness--completeness trade-off raises a fundamental question: does the systematic gap in field coverage arise from the model's inferential capacity, or is it an artifact of restricting inference to paper-only context while the ground truth was constructed from all three sources? To disentangle these factors, we evaluate Qwen3-30B-A3B-Instruct (the top-performing model on the main benchmark) under two conditions on the identical set of 100 cards used in Section~\ref{sec:humaneval}:

\begin{itemize}
  \item \textbf{Paper-only}: standard evaluation protocol, conditioned solely on paper text $P$, consistent with the main benchmark conditions.
  \item \textbf{All-source}: full multi-source context $P \cup S$, where $S$ comprises the GitHub README and Hugging Face artifact, constituting the same evidence available during ground-truth construction.
\end{itemize}

\paragraph{Results and Implications.}

\begin{table*}[t!]
\centering
\small
\begin{tabular}{l ccccc}
\toprule
& \textbf{Ann-1} & \textbf{Ann-2} & \textbf{Llama} & \textbf{GPT-OSS} & \textbf{Qwen} \\
\midrule
\multicolumn{6}{l}{\textit{Editor-Refined Benchmarks (Reference)}} \\
\quad Bench\_GPT-OSS        & 4.475    & 3.669    & 4.101    & 3.019    & 4.303 \\
\quad Bench\_Mistral        & 4.355    & 3.799    & 4.634    & 2.897    & 3.990 \\
\midrule
\multicolumn{6}{l}{\textit{Qwen3-30B-A3B-Instruct (Ablation)}} \\
\quad Paper-only            & 4.358    & 3.456    & 3.770    & 3.061    & 4.304 \\
\quad All-source            & 4.376    & 3.435    & 3.946    & 2.726    & 4.226 \\
\quad $\Delta$              & $+0.018$ & $-0.021$ & $+0.176$ & $-0.335$ & $-0.078$ \\
\midrule
Raw Baseline                & 3.709    & 2.966    & 3.130    & 2.951    & 3.447 \\
\bottomrule
\end{tabular}
\caption{%
  Ablation on source context. Qwen3-30B-A3B-Instruct is evaluated under paper-only and full multi-source conditions on the same 100-card test set used in Section~\ref{sec:humaneval}. Editor benchmarks and Raw Baseline are shown for reference. The near-zero $\Delta$ across reliable judges localises the completeness bottleneck to model inferential capacity rather than source sparsity.%
}
\label{tab:ablation_source}
\end{table*}

Table~\ref{tab:ablation_source} presents the mean Likert scores under both conditions alongside the editor-refined benchmarks for reference. Providing a full multi-source context yields only marginal and inconsistent variation: Ann-1 shifts by $+0.018$, Ann-2 by $-0.021$, and the Qwen judge by $-0.078$, with no judge registering a meaningful improvement attributable to the additional context. Crucially, the all-source condition continues to underperform both Bench\_GPT-OSS and Bench\_Mistral under domain annotators (4.376 vs.\ 4.475 / 4.355), despite enjoying equivalent evidential access during inference.

These results indicate that the completeness bottleneck is not an artifact of source sparsity in the evaluation setting, but is instead attributable to the model's long-context inferential capacity. In particular, the model struggles to locate, integrate, and populate sparse or implicitly stated metadata fields across heterogeneous multi-document inputs, consistent with documented limitations in dispersed-information retrieval from extended contexts. Consequently, the paper-only evaluation protocol constitutes a faithful and reproducible measure of model capability, rather than an artificially constrained one.

\subsection{Generator Divergence: Full Mean-Variance Analysis}
\label{sec:generator_divergence_full}

\begin{figure}[t]
    \centering
    \includegraphics[width=\linewidth]{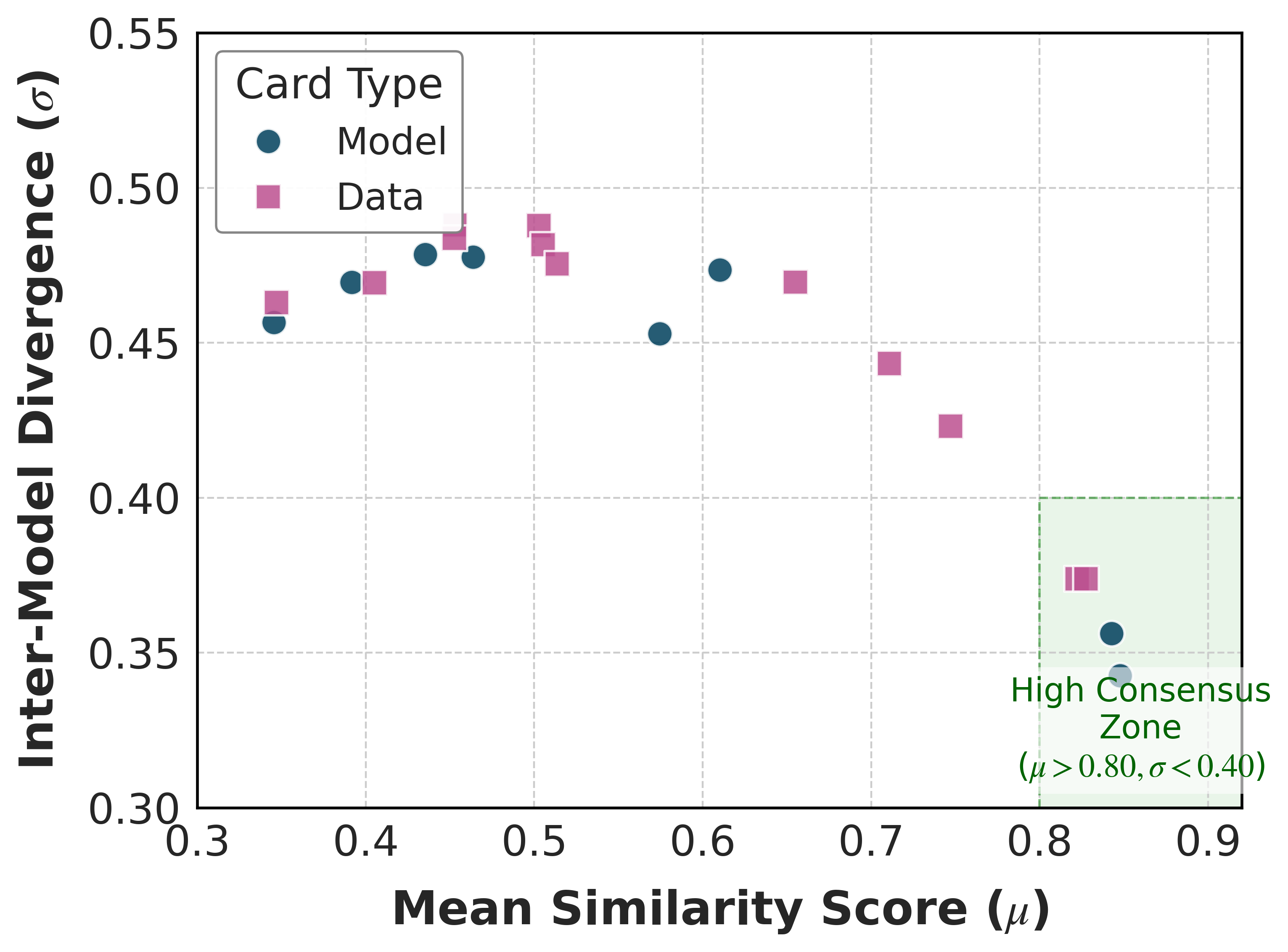}
    \caption{Mean-Variance analysis of semantic alignment. Mean BERTScore similarity ($\mu$, X-axis) versus variance (standard deviation $\sigma$, Y-axis) across card fields. High-consensus zones reflect evidence scarcity with converged placeholder responses.}
    \label{fig:scatter_consistency}
\end{figure}

Figure~\ref{fig:scatter_consistency} presents the full mean-variance distribution across all fields of cards. Fields such as \textit{Ethical Considerations} ($\mu = 0.848$) cluster in the high-consensus zone, predominantly reflecting evidence scarcity rather than genuine agreement: when documentation is absent from source materials, generators converge on similar placeholder responses (Appendix~\ref{sec:sparsity_analysis}). Fields with pronounced divergence ($\sigma > 0.4$) confirm that architecturally distinct generators surface complementary details from information-rich contexts, validating the ensemble design.

\subsection{Impact of Information Sparsity}
\label{sec:sparsity_analysis}

To explain the field-level performance stratification observed in Figure~\ref{fig:radar_charts}, we analyze the relationship between the density of the source text information and the generation performance using BERTScore as a diagnostic indicator of the difficulty of generation.

\begin{figure}[t]
    \centering
    \includegraphics[width=\linewidth]{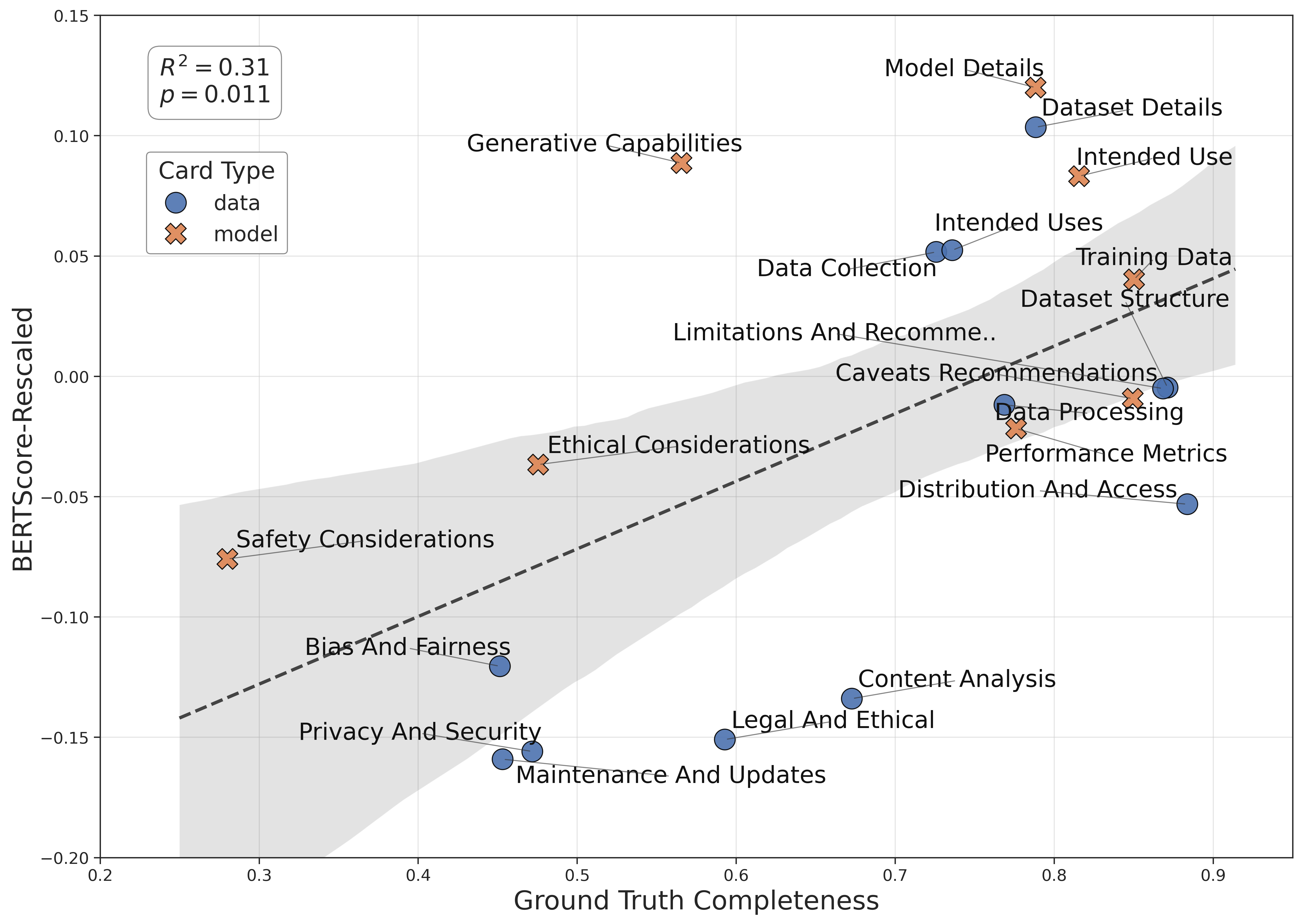}
    \caption{\textbf{Information Sparsity Impact on Generation Difficulty.} The relationship ($R^2=0.31, p=0.011$, OLS linear regression) between ground truth information density (X-axis, measured by completeness) and generation difficulty (Y-axis, measured by BERTScore as a diagnostic indicator). Fields with high sparsity (lower completeness) result in significantly lower generation success, highlighting the challenge of inferring abstract metadata from sparse signals.}
    \label{fig:sparsity_correlation}
\end{figure}

We hypothesize that the difficulty of generating abstract fields (\textit{Safety Considerations}, \textit{Bias And Fairness}) stems primarily from information sparsity in ground truth sources. High-density fields such as \textit{Model Details} provide abundant explicit signals, enabling high semantic alignment. Conversely, abstract fields contain sparse or implicit information, challenging models to infer unstated constraints without generating hallucinations.

Figure~\ref{fig:sparsity_correlation} demonstrates a statistically significant positive relationship ($R^2=0.31, p=0.011$) between the completeness of the ground truth (information density) and semantic similarity (BERTScore), indicating that 31\% of the variance of BERTScore is explained by the completeness of the source. This diagnostic analysis confirms that a primary bottleneck for current LLMs is the inferential capacity required to populate sparse metadata fields from weak textual signals, rather than simple retrieval failure.

\subsection{Comparative Lexical Analysis}
\label{sec:lexical_analysis}

\begin{figure*}[t]
    \centering
    \begin{minipage}{0.48\textwidth}
        \centering
        \includegraphics[width=\linewidth]{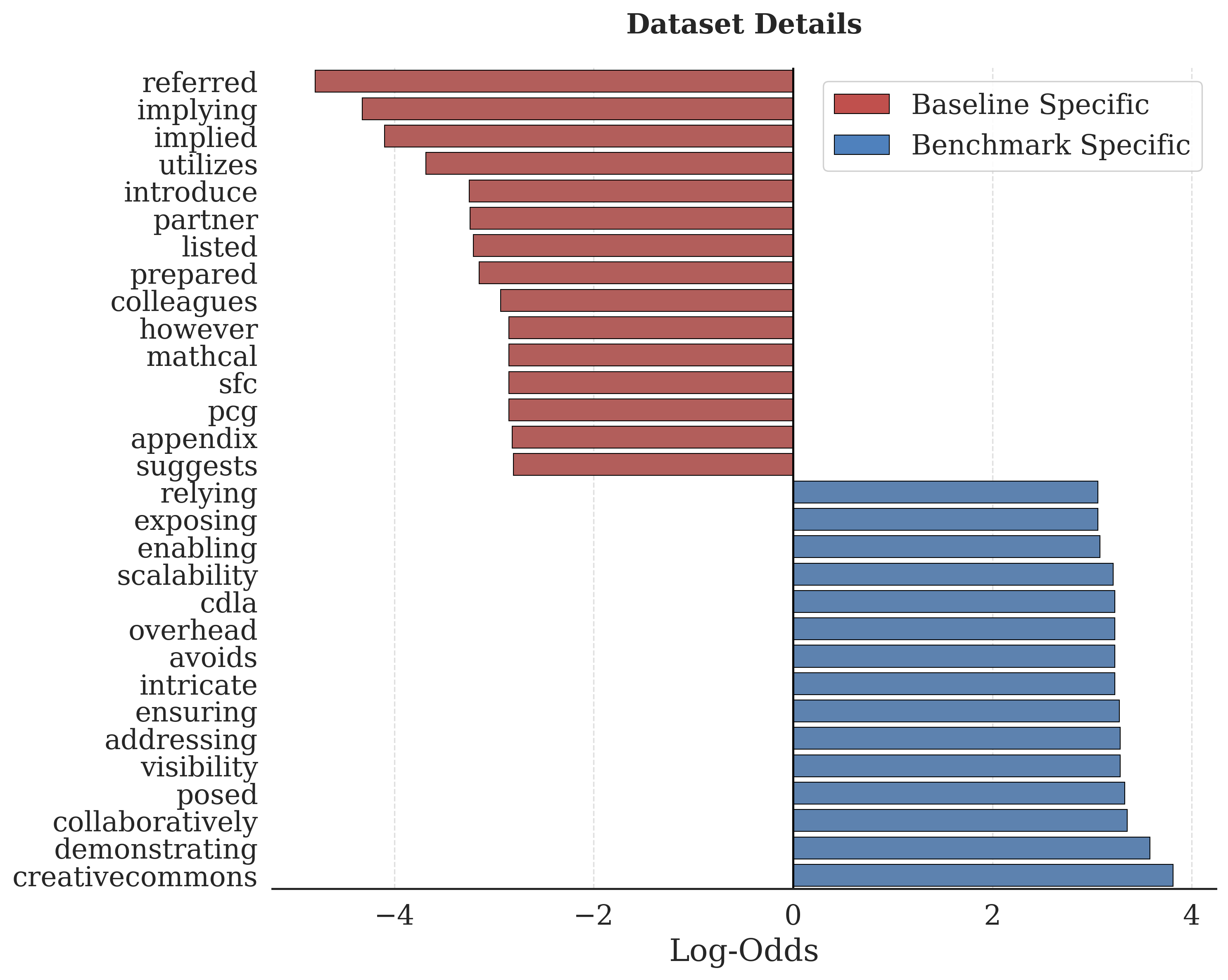}
        \caption*{(a) Dataset Details Divergence}
    \end{minipage}
    \hfill
    \begin{minipage}{0.48\textwidth}
        \centering
        \includegraphics[width=\linewidth]{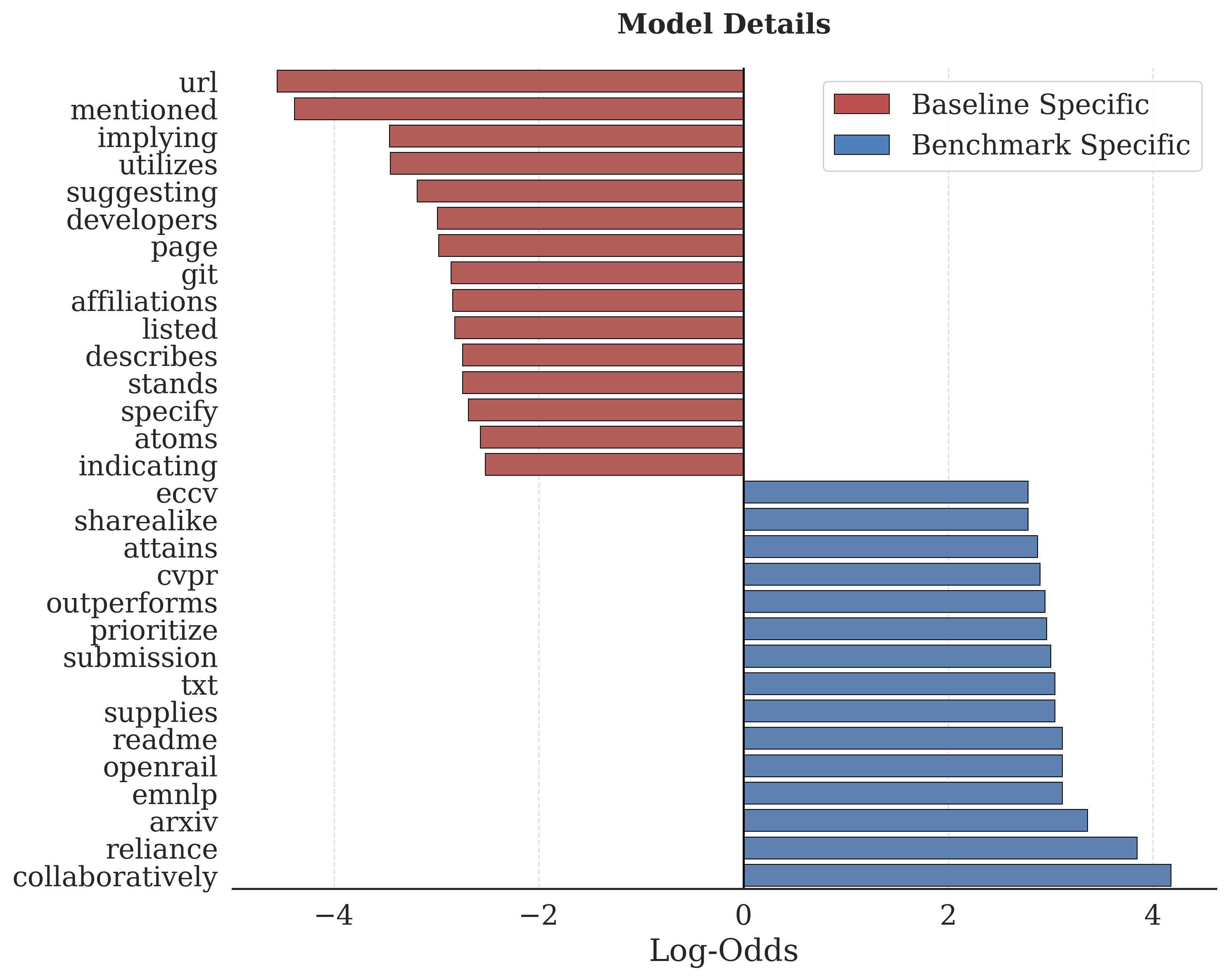}
        \caption*{(b) Model Details Divergence}
    \end{minipage}
    \caption{\textbf{Lexical Divergence (Log-Odds Ratio).} \textbf{Left (Red):} Words over-represented in Baselines, indicating narrative bias and format hallucinations. \textbf{Right (Blue):} Words specific to MetaGAI, highlighting the capture of high-value entities and technical specifications.}
    \label{fig:lexical_divergence}
\end{figure*}

We perform lexical analysis using the Log-Odds Ratio~\cite{monroe2008fightin} to assess the differences in linguistic quality between the generated content and ground truth. We focus on \textit{Model Details} and \textit{Dataset Details} fields, which require complex generation of open-ended, high-density technical information.

Figure~\ref{fig:lexical_divergence} reveals a systematic linguistic divergence. Baseline models (red) exhibit narrative-oriented patterns, frequently employing hedging language (\textit{suggests}, \textit{implying}) that summarizes experimental narratives rather than generating factual specifications. In contrast, MetaGAI ground truth (blue) uniquely captures high-value technical entities (\textit{CDLA}, \textit{OpenRAIL}) and precise licensing terms. This shows that our pipeline successfully generates concrete implementation specifications (artifact metadata) rather than paper narratives.

\subsection{Cost-Efficiency Analysis}
\label{appendix:cost_efficiency}

\begin{figure*}[t]
    \centering
    \begin{minipage}{0.48\linewidth}
        \centering
        \includegraphics[width=\linewidth]{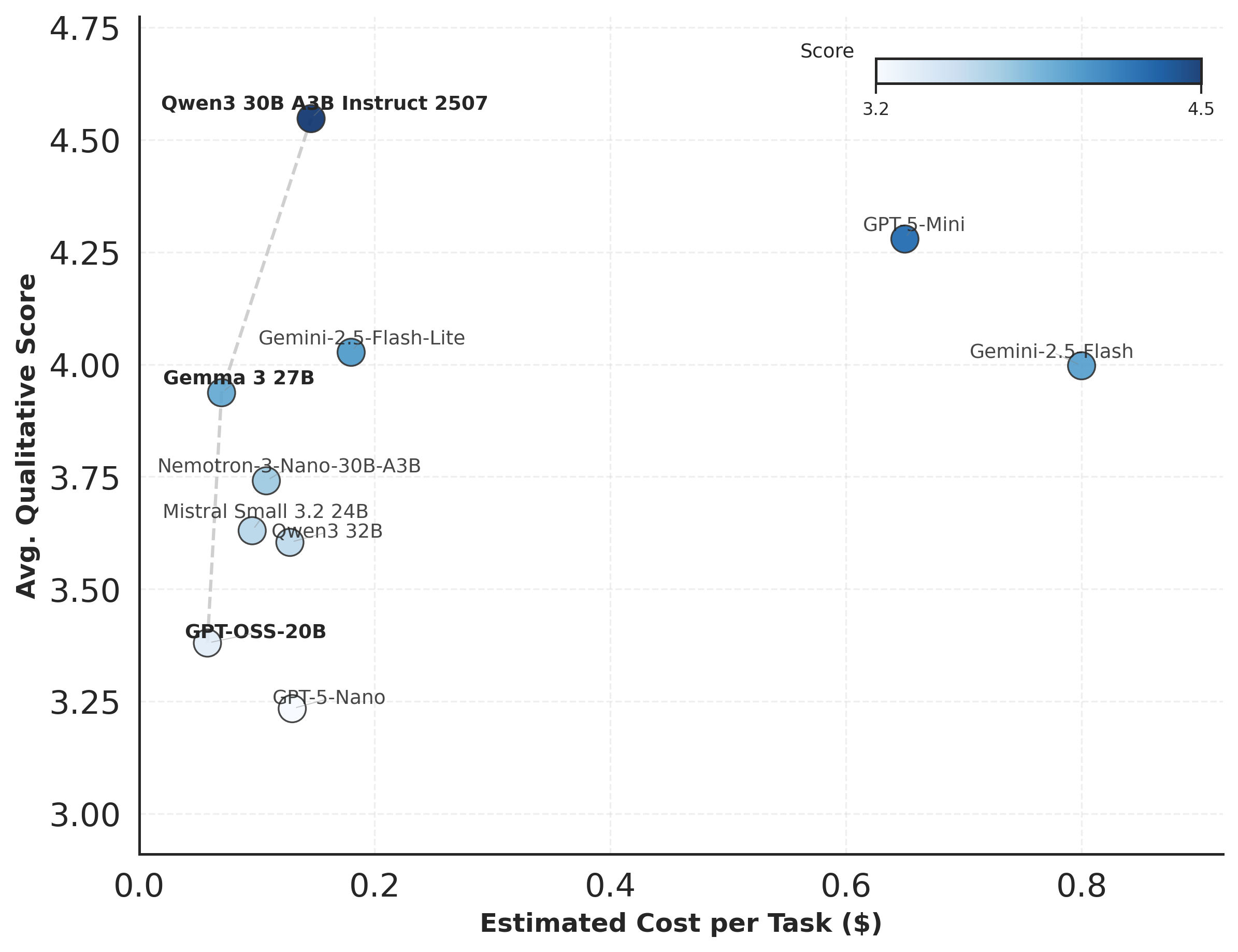}
        \caption*{\small (a) Model Card Generation}
    \end{minipage}
    \hfill
    \begin{minipage}{0.48\linewidth}
        \centering
        \includegraphics[width=\linewidth]{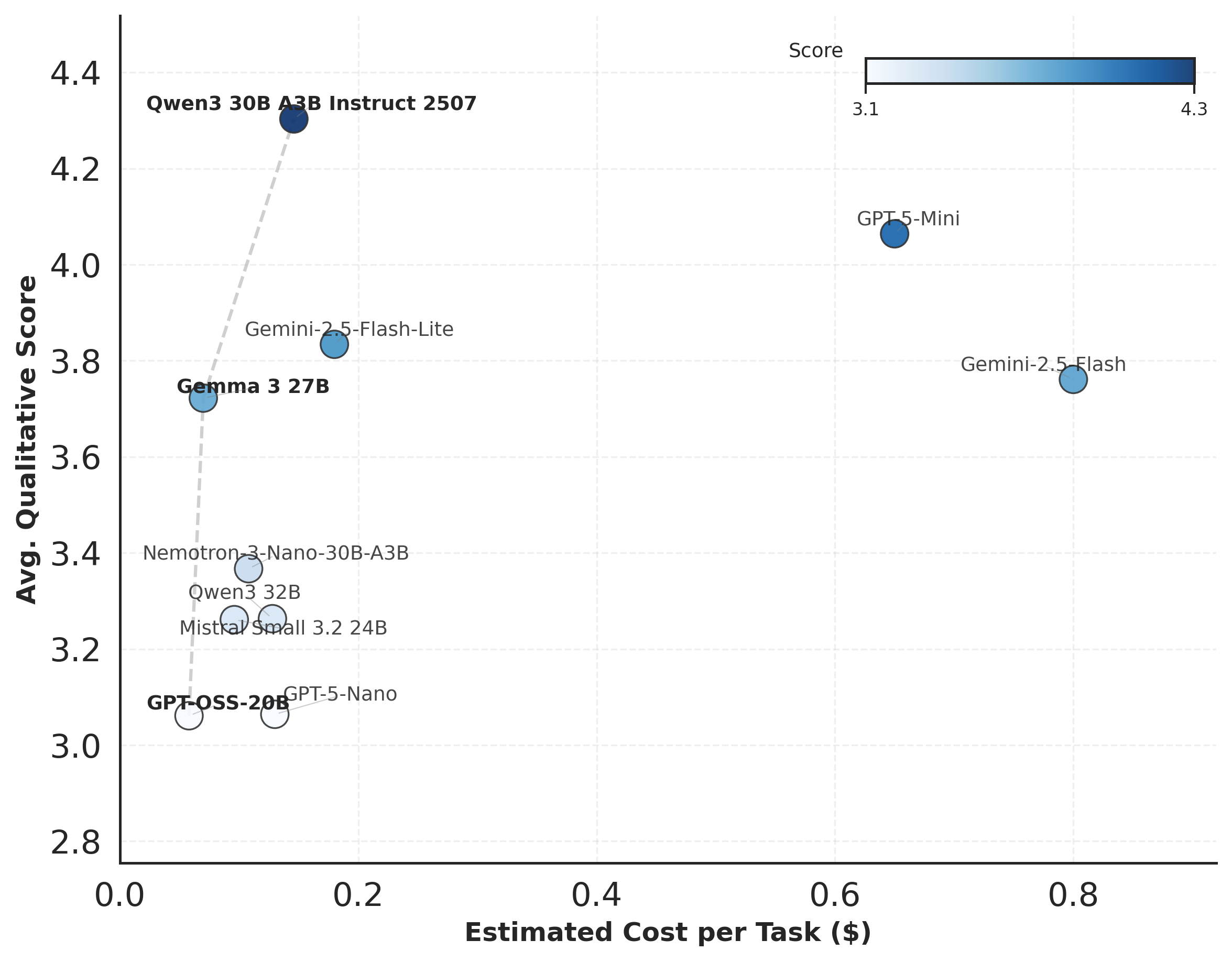}
        \caption*{\small (b) Data Card Generation}
    \end{minipage}
    \caption{\textbf{Cost-Efficiency Analysis.} The Pareto frontier (dashed grey line) highlights models that offer optimal quality for a given cost. Note the significant gap between the efficient open-weight frontier and closed-source proprietary models.}
    \label{fig:cost_efficiency}
\end{figure*}

To further evaluate the economic feasibility of deploying these models at the scale of millions of papers, we analyze the trade-off between generation quality and inference cost. We define a Cost Index normalized by standardizing token consumption (1M input / 0.2M output) based on OpenRouter pricing.

Figure~\ref{fig:cost_efficiency} visualizes this cost-quality landscape. The analysis reveals a distinct Pareto frontier dominated by open-weight architectures. Specifically, the sparse MoE model Qwen3-30B-A3B-Instruct achieves the highest qualitative scores while maintaining one of the lowest costs per task. In contrast, proprietary models such as Gemini-2.5-Flash and GPT-5-Mini, while capable, sit far to the right of the efficient frontier, incurring significantly higher costs (up to 5$\times$) without a proportional gain in generation fidelity. This disparity suggests that, for structured generation tasks, specialized open-weight models offer a far superior return on investment.

\section{Case Study}
\label{sec:case_studies}

To demonstrate the efficacy of the MetaGAI pipeline in complex retrieval and synthesis scenarios, we present two qualitative case studies illustrating how multi-source triangulation and multi-agent architecture mitigate common failure modes including information omission and incomplete evidence coverage.

\subsection{Multi-Source Triangulation}
\label{sec:case_bread}

We analyze the \textit{Model Architecture} generation for the paper ``Low-light Image Enhancement via Breaking Down the Darkness''~\cite{10.1007/s11263-022-01667-9}. As illustrated in Figure~\ref{fig:case_bread_triangulation}, high-fidelity output requires fusing distinct information modalities: the paper defines macro-level architectural topology, GitHub specifies implementation hyper parameters typically omitted from manuscripts, and Hugging Face validates entity alignment and deployment availability.

\begin{figure*}[t] 
    \centering
    \begin{tcolorbox}[
        colback=white, 
        colframe=teal!60!black, 
        title=\textbf{Triangulation Logic for ``Bread'' Architecture},
        fonttitle=\bfseries
    ]
        
        \scriptsize
        \begin{minipage}[t]{0.31\textwidth}
            \textbf{\textcolor{blue}{[Source] Paper}} \\
            \textit{``Figure 4 shows the overall architecture... comprises an illumination adjustment net (\textbf{IAN}), an adaptive noise suppression net (\textbf{ANSN}), and a color adaption net (\textbf{CAN}).''}
        \end{minipage}
        \hfill
        \vrule
        \hfill
        \begin{minipage}[t]{0.31\textwidth}
            \textbf{\textcolor{purple}{[Source] GitHub}} \\
            \textit{(Code Implementation)} \\
            \texttt{class IAN(nn.Module):} \\
            \texttt{  layers = [Conv3x3] * 3} \\
            \texttt{  \# 3 down-sampling} \\ 
            \texttt{  \# 3 up-sampling} \\   
            \texttt{  out = torch.sigmoid(x)}
        \end{minipage}
        \hfill
        \vrule
        \hfill
        \begin{minipage}[t]{0.31\textwidth}
            \textbf{\textcolor{orange!80!black}{[Source] Hugging Face}} \\
            \textit{(Metadata Verification)} \\
            \texttt{Task: Image Restoration} \\
            \texttt{Tags: [PyTorch, bread]} \\
            \textit{Confirms model identity and availability.}
        \end{minipage}

        \vspace{6pt}
        \hrule
        \vspace{6pt}

        \normalsize
        \textbf{MetaGAI Editor Decision (GPT-OSS-20B)} \\
        \small
        \textbf{Selected Candidate:} Qwen2.5-7B (Draft B) \\
        \textbf{Source Types Verified:} Paper, GitHub, Hugging Face \\
        \textbf{Reasoning:} \textit{``Candidate B accurately captures the proposed Bread architecture, whereas the other candidates lack relevant details.''}

        \vspace{4pt}
        \begin{tcolorbox}[colback=gray!5, colframe=gray!20, sharp corners, boxrule=0.5pt]
            \textbf{Final Generated Content:} \\
            \small
            \textbf{Bread} is a three-stage CNN pipeline:
            \begin{enumerate}[leftmargin=1.5em, itemsep=0pt, topsep=0pt, partopsep=0pt]
                \item \textbf{Illumination Adjustment Network (IAN)} predicts a relative illumination map to brighten the luminance.
                \item \textbf{Adaptive Noise Suppression Network (ANSN)} denoises the brightened luminance using multiple suppression strengths fused by a Noise Fusion Module.
                \item \textbf{Color Adaption Network (CAN)} refines chrominance guided by the denoised luminance.
            \end{enumerate}
            \textbf{\textit{Synthesis:}} Each sub-network follows a symmetric encoder–decoder with 3 down-sampling and 3 up-sampling layers, 3×3 convolutions, and a sigmoid output (except ANSN).
        \end{tcolorbox}
    \end{tcolorbox}
    \caption{\textbf{Multi-Source Triangulation.} The Editor Agent synthesizes the high-level topology from the Paper, specific layer depths and activation functions from GitHub, and validates alignment via Hugging Face, resulting in a complete specification.}
    \label{fig:case_bread_triangulation}
\end{figure*}

This case validates the pipeline's ability to resolve the granularity gap between conceptual descriptions and implementation specifications. While the paper provides the architectural framework, the GitHub repository provides the concrete hyper parameters necessary for reproducibility. The Editor Agent (GPT-OSS-20B) correctly identified that Candidate B integrated evidence across all three sources, selecting it over simpler drafts that merely paraphrased the abstract.

\subsection{Editor-Driven Information Synthesis}
\label{sec:case_synthesis}

While retrieval ensures evidence access, individual Generator Agents frequently exhibit narrow focus, generating information from specific paper sections while overlooking complementary content. The Editor Agent provides critical synthesis capabilities to resolve this incompleteness.

Continuing with the same paper, we analyze the \textit{Robustness Metrics} field for the ``Bread'' model. As shown in Figure~\ref{fig:case_editor_synthesis}, the two Generator Agents produced factually accurate but incomplete drafts: Candidate A (Olmo3-7B) generated only comparative benchmark results, while Candidate B (Qwen2.5-7B) focused exclusively on the findings of the ablation study. The Editor Agent (GPT-OSS-20B) recognized that both candidates captured orthogonal robustness dimensions: external performance validation versus internal architectural stability. Rather than selecting a single candidate, the editor merged non-redundant information to generate comprehensive coverage addressing both comparative benchmarking and component ablation analysis. This synthesis capability demonstrates a critical advantage of our multi-agent architecture over single-model generation approaches, which typically exhibit selection bias toward particular evidence types.

\begin{figure*}[t] 
    \centering
    \begin{tcolorbox}[
        colback=white, 
        colframe=teal!60!black, 
        title=\textbf{Synthesis Logic for ``Bread'' Robustness},
        fonttitle=\bfseries
    ]
        
        \scriptsize
        \begin{minipage}[t]{0.47\textwidth}
            \textbf{\textcolor{blue}{Candidate A (Olmo3-7B)}} \\
            \textit{(Focus: External Benchmarks)} \\
            \textit{``Bread achieves the lowest \textbf{NIQE and LOE scores} on the DICM, NPE, and VV datasets, outperforming all competing methods.''} \\
            \vspace{2pt}
            \textcolor{red!70!black}{\xmark \ \textit{Missing internal analysis.}}
        \end{minipage}
        \hfill
        \vrule
        \hfill
        \begin{minipage}[t]{0.47\textwidth}
            \textbf{\textcolor{purple}{Candidate B (Qwen2.5-7B)}} \\
            \textit{(Focus: Internal Ablation)} \\
            \textit{``\textbf{Ablation studies} show that removing key modules such as the noise fusion module causes marked drops in PSNR/SSIM.''} \\
            \vspace{2pt}
            \textcolor{red!70!black}{\xmark \ \textit{Missing comparative stats.}}
        \end{minipage}

        \vspace{6pt}
        \hrule
        \vspace{6pt}

        \normalsize
        \textbf{MetaGAI Editor Decision (GPT-OSS-20B)} \\
        \small
        \textbf{Action:} Merge Non-Redundant Information \\
        \textbf{Reasoning:} \textit{``Both candidates describe valid, orthogonal aspects of robustness. Candidate A covers external metrics (NIQE/LOE), while Candidate B covers internal validity (Ablation). Merging creates a holistic summary.''}

        \vspace{4pt}
        \begin{tcolorbox}[colback=gray!5, colframe=gray!20, sharp corners, boxrule=0.5pt]
            \textbf{Final Generated Content:} \\
            \small
            \begin{itemize}[leftmargin=1.5em, itemsep=0pt, topsep=0pt]
                \item \textbf{External Performance:} Bread achieves the lowest NIQE and LOE scores on the DICM, NPE, and VV datasets.
                \item \textbf{Internal Stability:} Ablation studies confirm framework robustness; removing the noise fusion module causes marked drops in PSNR/SSIM.
            \end{itemize}
        \end{tcolorbox}
    \end{tcolorbox}
    \caption{\textbf{Resolving Incompleteness via Synthesis.} The Editor Agent detects that Candidates A and B focus on orthogonal aspects of robustness (External Benchmarks vs. Internal Ablation). Instead of selecting a single winner, the Editor merges them to generate a comprehensive entry.}
    \label{fig:case_editor_synthesis}
\end{figure*}

\end{document}